\documentclass[submit]{theme/bioRxiv}

\usepackage{tikz}
\usepackage{xcolor}
\usepackage{threeparttable}
\newcommand*\circled[1]{%
  \tikz[baseline=(char.base)]{
    \node[shape=circle,fill=black,inner sep=1pt,minimum size=1.5em,text=white] (char) {#1};
  }
}

\usepackage{glossaries}
\newacronym[
    shortplural={RQs}, 
    longplural={Research Questions}
]{rq}{RQ}{Research Question}

\newacronym{ifc}{IFC}{Imaging Flow Cytometry}
\newacronym[
    shortplural={iPSCs},
    longplural={Induced Pluripotent Stem Cells}
]{ipsc}{iPSC}{Induced Pluripotent Stem Cell}
\newacronym{rbc}{RBC}{Red Blood Cells}
\newacronym{fc}{FC}{Flow Cytometry}
\newacronym{isf}{ISF}{Immunological Synapse Formation}

\newacronym{ccd}{CCD}{Charge-Coupled Device}
\newacronym{cmos}{CMOS}{Complementary Metal-Oxide-Semiconductor}
\newacronym{tdi}{TDI}{Time-Delay Integration}
\newacronym{ebv}{EBv}{Epstein-Barr virus}
\newacronym{snr}{SNR}{Signal-to-noise ratio}
\newacronym{sea}{SEA}{Staphylococcus aureus enterotoxin A}

\newacronym{ai}{AI}{Artificial Intelligence}
\newacronym{ml}{ML}{Machine Learning}
\newacronym{dl}{DL}{Deep Learning}
\newacronym{bt}{BT}{Barlow Twins}
\newacronym{lr}{LR}{Learning Rate}
\newacronym[
    shortplural={MLPs},
    longplural={Multilayer Perceptrons}
]{mlp}{MLP}{Multilayer Perceptron}
\newacronym[
    shortplural={SVMs},
    longplural={Support Vector Machines}
]{svm}{SVM}{Support Vector Machine}

\newacronym[
    shortplural={MAEs},
    longplural={Masked Autoencoders}
]{mae}{MAE}{Masked Autoencoder}

\newacronym[
    shortplural={CNNs},
    longplural={Convolutional Neural Networks}
]{cnn}{CNN}{Convolutional Neural Network}
\newacronym{ssl}{SSL}{Self-Supervised Learning}

\newacronym{resnet}{ResNet}{Residual Network}
\newacronym[
    shortplural={ViTs},
    longplural={Vision Transformers}
]{vit}{ViT}{Vision Transformer}
\newacronym{cls}{CLS}{Classification Token}
\newacronym{flops}{FLOPs}{Floating Point Operations per Second}

\newacronym{cpc}{CPC}{Contrastive Predictive Coding}
\newacronym{moco}{MoCo}{Momentum Contrast}
\newacronym{byol}{BYOL}{Bootstrap Your Own Latent}
\newacronym{swav}{SwAV}{Swapping Assignments Between Multiple Views}

\newacronym{ari}{ARI}{Adjusted Rand Index}
\newacronym{nmi}{NMI}{Normalized Mutual Information}
\newacronym{ema}{EMA}{Exponential Moving Average}
\newacronym{sgd}{SGD}{Stochastic Gradient Descent}
\newacronym{bn}{BN}{Batch Normalization}
\newacronym{ln}{LN}{Layer Normalization}
\newacronym{umap}{UMAP}{Uniform Manifold Approximation and Projection}
\newacronym{pca}{PCA}{Principal Component Analysis}
\newacronym{t-sne}{t-SNE}{t-distributed Stochastic Neighbor Embedding}
\newacronym{knn}{kNN}{k-Nearest Neighbor}

\newacronym{auroc}{AUROC}{Area Under the Receiver Operating Characteristic Curve}
\newacronym{roc}{ROC}{Receiver Operating Characteristic}

\newacronym{tp}{TP}{True Positive}
\newacronym{fp}{FP}{False Positive}
\newacronym{tn}{TN}{True Negative}
\newacronym{fn}{FN}{False Negative}
\newacronym{pp}{pp}{percentage points}

\newglossaryentry{backpropagation}{
    name=Backpropagation, 
    text=backpropagation, 
    description={Propagating gradients obtained from an error function in backward direction through a network}
}

\usepackage{lipsum}

\usepackage{multirow} 
\usepackage{booktabs}

\begin{document}

 \leadauthor{Yamachui Sitcheu}

\title{IFC-SSL-Bench: Benchmarking Self-Supervised Learning for Annotation-Efficient Multispectral Imaging
Flow Cytometry}
\shorttitle{IFC-SSL-Bench}

\author[1,\Letter]{Angelo Jovin Yamachui Sitcheu \orcidlink{0009-0002-9384-6496}}
\author[1]{Roxana-Teodora Fetescu}
\author[1]{Thomas Lautenschlager \orcidlink{0009-0002-3214-9363}}
\author[1,2]{Nils Friederich \orcidlink{0009-0003-9377-3140}}
\author[1]{Linus Schilpp}
\author[1]{Stephan Allgeier \orcidlink{0000-0002-1218-448X}}
\author[1,3]{Luca Deininger \orcidlink{0009-0005-5125-6422}}
\author[4,5,6,7,8]{Daniel Hübschmann \orcidlink{0000-0002-6041-7049}}
\author[1]{Ralf Mikut \orcidlink{0000-0001-9100-5496}}

\affil[1]{Institute for Automation and Applied Informatics, Karlsruhe Institute of Technology, Eggenstein-Leopoldshafen, Germany}
\affil[2]{Institute of Biological and Chemical Systems, Karlsruhe Institute of Technology, Eggenstein-Leopoldshafen, Germany}
\affil[3]{Department of Biosystems Science and Engineering, ETH Zürich, Basel, Switzerland}
\affil[4]{Heidelberg Institute for Stem Cell Technology and Experimental Medicine, HI-STEM gGmbH, Heidelberg, Germany}
\affil[5]{Innovation and Service Unit for Bioinformatics and Precision Medicine, German Cancer Research Center, Heidelberg, Germany}
\affil[6]{German Cancer Consortium, DKTK, Heidelberg, Germany}
\affil[7]{Division of Translational Precision Medicine, Institute of Human Genetics, Heidelberg University, Heidelberg, Germany}
\affil[8]{Computational Oncology Group, Molecular Precision Oncology Program, National Center for Tumor Diseases Heidelberg and German Cancer Research Center, Heidelberg, Germany}
\date{}

\maketitle

\begin{abstract}
Imaging Flow Cytometry (IFC) generates large multispectral datasets, yet the need for expert annotations bottlenecks the development of supervised deep learning models for cellular image analysis. While Self-Supervised Learning (SSL) offers a solution, its application to IFC remains fragmented and lacks systematic benchmarking. To address this, we present IFC-SSL-Bench, a modular framework for evaluating SSL in IFC. We benchmark six SSL approaches across four backbones on two heterogeneous datasets. By assessing downstream classification performance across various annotation regimes of the training sets ($1\%\mbox{--}100\%$), we quantify annotation efficiency and identify optimal configurations for high-dimensional biological imaging. We demonstrate that SSL pretraining reduces annotation dependency. SSL matches the performance of fully supervised baselines trained on $100\%$ of annotated data while using as little as $5\%$ of annotations for morphological classification, with performance advantages in low data regimes under $5\%$ of annotated data. Across both datasets, the contrastive approach MoCo-v2  paired with ConvNeXt backbones emerged as the top-performing configuration, with ConvNeXtV1\mbox{--}T performing best for immunological synapse classification and ConvNeXtV2\mbox{--}T for morphological classification. We identify a clear pattern in transfer dynamics: ImageNet initialization benefits morphological classification, consistent with its visual similarity to natural images, whereas pretraining from random initialization better captures the more domain-specific immunological synapse classification. SSL elevates class-specific F1-scores by over $35$ percentage points for structurally complex classes at $1\%$ of annotated data. We synthesize these empirical insights into the first unified set of guidelines for deploying self-supervised workflows in high-throughput IFC, establishing a benchmark and resource suite that lowers the annotation barrier for future work in IFC.
\end{abstract}


\begin{corrauthor}
angelo.sitcheu\at kit.edu
\end{corrauthor}

\section*{Introduction}\label{sec:introduction}
\gls{ifc} is a modern technology that combines the quantitative, high-throughput capabilities of \gls{fc}  with the spatial and morphological detail provided by microscopy~\citep{basiji2007cellular}. In practice, this is achieved by hydrodynamically focusing cells into a single-file stream that passes through a laser-illuminated imaging zone. High-speed detectors, synchronized with the cell flow, capture multispectral images using combinations of brightfield, darkfield, phase-contrast, and fluorescence channels, to achieve sharp subcellular resolution at high velocities~\citep{rees_imaging_2022, ugawa_recent_2024}. By capturing multiple images of each cell as it passes through the imaging region, \gls{ifc} produces high-dimensional image datasets that provide unprecedented single-cell depth~\citep {ugawa_recent_2024}. This hybrid technique has advanced research in immunology, oncology, cell biology, and environmental science by allowing detailed examination of heterogeneous cell populations and subtle phenotypic variations with robust statistical confidence~\citep{rees_imaging_2022, dunker2021pollen, dunker2022potential, vonficht2025ultra}.

Despite its powerful capabilities, \gls{ifc} faces significant challenges that stem not only from the inherent biological heterogeneity of cell populations~\citep{10.3389/fimmu.2021.787574}, but also from the technical constraints, and therefore hinders effective data analysis. The primary technical hurdle in \gls{ifc} is the fundamental trade-off to set between throughput and image resolution. While increasing magnification or exposure time gradually improves spatial resolution and fine morphological detail, it inherently reduces acquisition speed. Conversely, high-throughput demands shorter exposure times, which results in reduced image quality with lower magnification and/or low \gls{snr}~\citep{hua2024light}. In multichannel acquisitions, fluorescence signals are additionally subject to background and to spectral crosstalk between detectors, which can obscure weak markers and distort the measured intensity distributions~\citep{mckinnon2018flow}. The appearance of a given cell type therefore depends not only on its biology but also on how it was acquired, so that the morphological differences of interest are superimposed on variation of purely technical origin.

Beyond these acquisition challenges, the analysis of \gls{ifc} data also faces various bottlenecks. Traditionally, \gls{ifc} data analysis has relied on manual gating or software-assisted pipelines using handcrafted feature extraction. Tools such as CellProfiler~\citep{carpenter2006cellprofiler} improved throughput by automating cell feature extraction of hundreds of morphological and intensity-based descriptors. However, these classical approaches still require extensive domain expertise to select and tune relevant features for specific biological assays, making the analysis labor-intensive and susceptible to operator subjectivity~\citep{stirling2021cellprofiler, julian2023machine}. This reliance on predefined features introduces variability and limits scalability, as manual or semi-automated gating is not only time-consuming but also prone to inconsistencies across different operators~\citep{rees_imaging_2022, https://doi.org/10.1002/cyto.a.23794}.

With the advent of \gls{dl}, many of these challenges have been partially mitigated. \gls{dl} models can automatically learn complex features from raw data, reducing the dependence on manual feature engineering. This transition from manual feature engineering to \gls{dl}-based image analysis has also occurred in the field of \gls{ifc}~\citep{DBLP:conf/fedcsis/KrajsicHD23, Kranich01012020, delikoyun2025realtimedeeplearningphase}. Several specialized pipelines have emerged to facilitate this transition, including PXPermute~\citep{boushehri2024pxpermute} for feature importance analysis, Deepometry~\citep{doan2021deepometry} for multi-channel phenotyping, Deep Cytometry~\citep{li2019deep} for real-time sorting applications, and Small U-Net~\citep{kaliman2025small} for high-throughput segmentation. 

Despite these successes, the models still require large amounts of accurately annotated data for supervised training \citep{HONG2025111992, wang2021review}. Given the high cost and expertise required to annotate images, obtaining sufficiently large and diverse annotated datasets remains a significant bottleneck, leading to scarce datasets, limiting generalization performance, and creating a need for annotation-efficient strategies~\citep{HONG2025111992}. To address this, \gls{ssl} methods have emerged as a promising alternative \citep{zeng2024self, rani2024self}. \gls{ssl} leverages unannotated data to learn useful representations, reducing the need for costly manual annotations. \gls{ssl} is a paradigm well suited to \gls{ifc}, which naturally generates vast amounts of unannotated cellular images at high-throughput scales. The robust performance of \gls{ssl} in processing high-dimensional data, most notably in natural language \citep{min2023recent} and video analysis \citep{schiappa2023self}, has provided a strong impetus for its application to other fields such as transcriptomics~\citep{DBLP:conf/icml/OvcharenkoBTDVS25, richter2025delineating} and visual geo-localization~\citep{xiao2025vg}. In medical imaging, \gls{ssl} has reached a high level of maturity: in digital histopathology, for example, the \gls{ssl} method \gls{bt} was used to pretrain a model on whole slide images and demonstrated remarkable robustness to cohorts variations and tissue types~\citep{liu2025self}. Similarly, on widely used benchmarks like MedMNIST~\citep{yang_medmnist_2023}, \gls{ssl} has been investigated to provide clear domain guidelines. Bundele et al.~\citep{bundele2024evaluating} benchmark eight \gls{ssl} methods across eleven MedMNIST datasets, varying annotation fraction and initialization, and find that ImageNet initialization consistently benefits performance, especially at low annotation fractions. Boukdir similarly reports transferable \gls{ssl} representations across diagnostic domains~\citep{boukdir2026empirical}. More recently in radiology with nnFoundation, self-supervised pretraining at scale has produced general purpose encoders that transfer across more than a hundred downstream tasks~\citep{harsy2026nnfoundation}.

Cellular imaging has followed the same trajectory. On Cell Painting, Kraus et al. trained \glspl{mae} on millions of images and recovered known relationships between genetic and chemical perturbations~\citep{kraus2024masked}, while Kim et al. compared self-supervised objectives directly on more than $10,000$ compounds and found that DINO outperformed \gls{mae}, contrastive learning and the engineered CellProfiler pipeline~\citep{kim2025self}. Self-distillation has since been scaled across assays and cell types: Moutakanni et al. trained DINOv2 on several million multichannel images from the Human Protein Atlas and Cell Painting, matching supervised performance on protein localization tasks with $1\%$ of the annotations~\citep{moutakanni2025cell}, and Zedda et al. assembled $1.25$ million single-cell images from $18$ datasets to build a dedicated foundation model for red blood cell morphology~\citep{ZedLuc_RedDino_MICCAI2025}.

However, these successes do not seamlessly translate to \gls{ifc}, which presents a unique set of challenges. Current literature focuses on isolated methods, demonstrating, for example, that a specific method like MoCo-v2~\citep{DBLP:journals/corr/abs-2003-04297} can be used for leukocyte differential analyses~\citep{Huang2026-if}, or \gls{bt}~\citep{DBLP:conf/icml/ZbontarJMLD21} for investigating immunological synapse formation and characterizing antibodies~\citep{shetab_boushehri_explainable_2023}. These studies fail to address whether these choices are optimal, leaving the comparative potential of other paradigms unexplored.

Collectively, these limitations underscore a fundamental gap in the principled evaluation of \gls{ssl} approaches for \gls{ifc}, raising the following key research question: How can \gls{ssl} be effectively applied and optimized for multispectral \gls{ifc} data?

This central inquiry is decomposed into two specific research questions (RQs):
\begin{itemize}
    \item \textbf{RQ1: Methodological and architectural effectiveness}: Which state-of-the-art \gls{ssl} methods and backbone architectures are most effective for learning reliable and discriminative embeddings from multispectral \gls{ifc} data?
    \item \textbf{RQ2: Annotation efficiency}: How does \gls{ssl} reduce annotation needs on  downstream classification?
    
\end{itemize}
 By addressing these questions, this work aims to provide a comprehensive benchmark for \gls{ssl} in the \gls{ifc} domain, offering insights into the optimal combination of algorithms, architectures, and preprocessing strategies for high-dimensional biological imaging. Our primary contributions are:
\begin{enumerate}
\item \textbf{A unified benchmarking pipeline for multispectral \gls{ifc}}: We introduce \gls{ifc}-\gls{ssl}-Bench, a standardized, modular, reproducible and extensible framework that unifies data processing, pretraining, and downstream evaluation to ensure fair comparisons across diverse \gls{ssl} methods for \gls{ifc}.
\item \textbf{Comprehensive empirical analysis}: We systematically evaluate six state-of-the-art invariance-based \gls{ssl} methods across four backbone architectures on two distinct \gls{ssl} datasets, providing the first structured comparison of self-supervised representation learning efficacy in this domain, identifying superior performance regimes.
\item \textbf{Annotation-efficient learning}: We deliver empirical evidence on the utility of \gls{ssl} under annotation scarcity. By quantifying performance gains across varying fractions of annotated data, we provide insights in optimizing \gls{ssl} in annotation-constrained scenarios.
\item \textbf{Actionable guidelines \& open resource}: We derive best-practice recommendations for models, \gls{ssl} methods, and initialization strategies tailored to \gls{ifc} data and release our full codebase under an open license at \url{https://github.com/hciays/IFC-SSL-Bench} to foster reproducibility and advance future research in \gls{ssl} applications for \gls{ifc}.
\end{enumerate}

\section*{Results}\label{sec:results}
To address the lack of systematic evaluation for \gls{ssl} in \gls{ifc}, we developed a benchmarking framework designed to decouple the influence of \gls{ssl} paradigms, backbone architectural choices, and domain-specific knowledge.

\subsection*{The Benchmarking Pipeline}\label{subsec:meth_pipeline}

\begin{figure*}[htb!]
\centerline{\includegraphics[width=\linewidth]{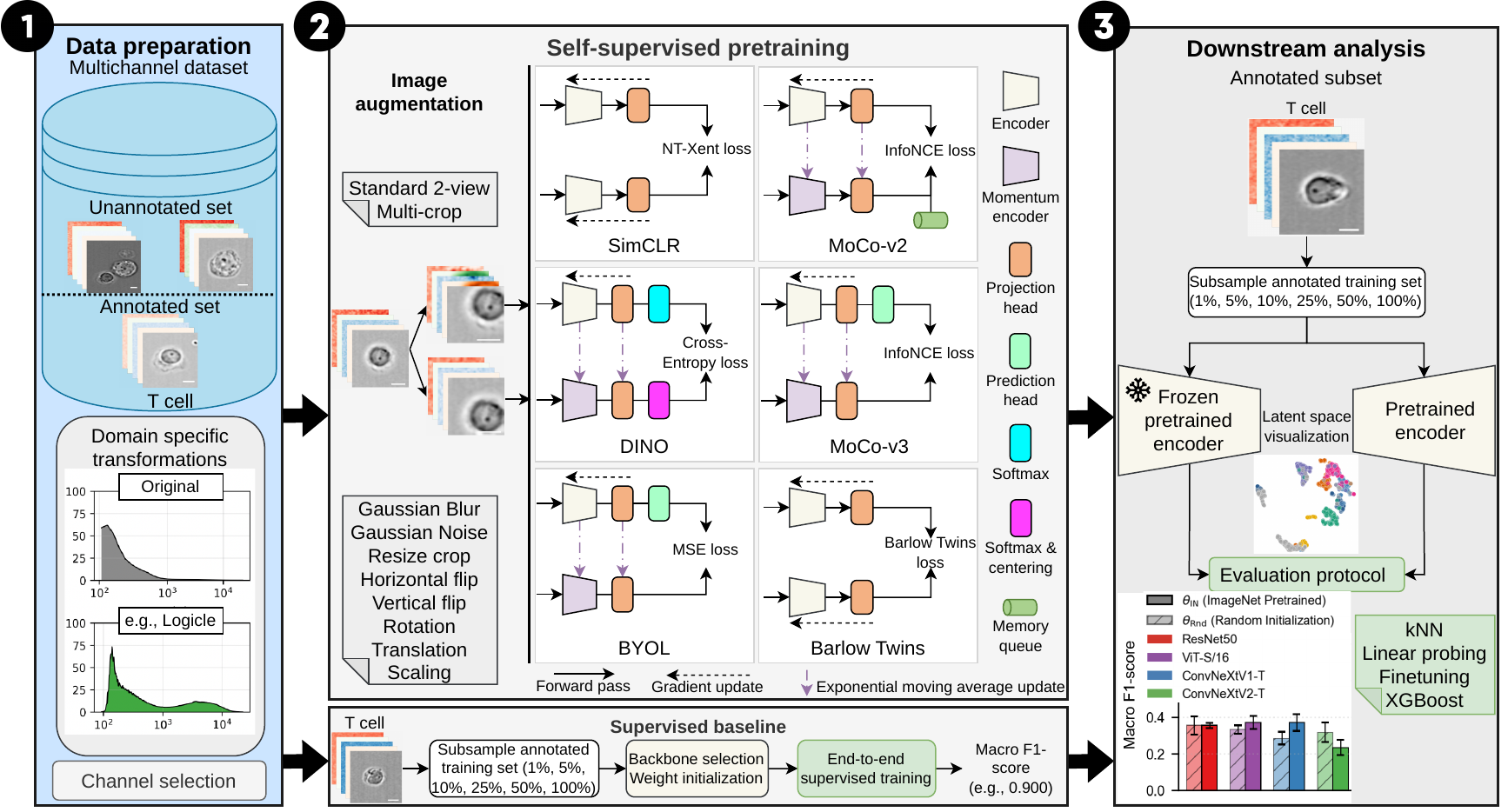}}
\caption{Overview of the modular \gls{ssl} benchmarking framework. The pipeline is organized into three core modules: 1, data preparation. This module covers the dataset split generation, channel selection, and domain-specific transformations. 2, the learning module, which consists of the self-supervised pretraining and supervised baseline. Six self-supervised objectives spanning three families of invariance-based \gls{ssl} are pretrained on the unannotated set: contrastive (SimCLR, MoCo-v2, MoCo-v3), asymmetric (\acrshort{byol}, DINO), and redundancy reduction (\gls{bt}). In parallel, the annotated set is used to train end-to-end supervised baselines for each backbone, weight initialization, and annotated data fraction ($1\%$, $5\%$, $10\%$, $25\%$, $50\%$, and $100\%$). 3, downstream analysis evaluates the pretrained encoders on image classification at the same annotated data fractions as the supervised baselines through four evaluation protocols (\acrshort{knn}, linear probing, XGBoost, and full finetuning) quantified by macro F1 scores. This quantitative analysis is complemented by a qualitative assessment of the learned representations. Scale bars in the multichannel sample images represent $5\mu m$.\label{fig:method}}
\end{figure*}

\autoref{fig:method} describes our proposed modular framework for benchmarking \gls{ssl} methods on \gls{ifc}. The pipeline standardizes \gls{ssl} evaluation through three integrated stages: \circled{1}, a data preparation module (Methods, Data Preparation) performs dataset splitting into an unannotated set $\mathcal{D}_{\mathrm{una}}$ for pretraining and an annotated set $\mathcal{D}_{\mathrm{ann}}$ for downstream analysis, channel selection, and domain-specific transformations. This is followed by \circled{2}, a flexible pretraining module (Methods, Self-Supervised Pretraining) designed to decouple the effects of model architecture, initialization strategy, and \gls{ssl} methodology. On the self-supervised path, six objectives spanning contrastive, asymmetric, and redundancy reduction families are pretrained on $\mathcal{D}_{\mathrm{una}}$; all of them draw their views from the same augmentation operators and parameters for a given backbone, while the number and spatial scale of views follow each method's original formulation, since the multi-crop scheme of DINO is intrinsic to its objective rather than a preprocessing choice. In parallel, fully supervised models are trained end-to-end on $\mathcal{D}_{\mathrm{ann}}$ for each combination of backbone, weights initialization choice, and fraction of annotated data. The self-supervised path therefore isolates the contribution of the objective, and the supervised path provides the reference against which annotation efficiency and representation quality are measured. Finally \circled{3}, an evaluation module (Methods, Downstream Analysis) responsible for downstream analyses, integrating both quantitative performance metrics and qualitative representation assessments.

\begin{figure*}[!htb]
\centerline{\includegraphics[width=\linewidth]{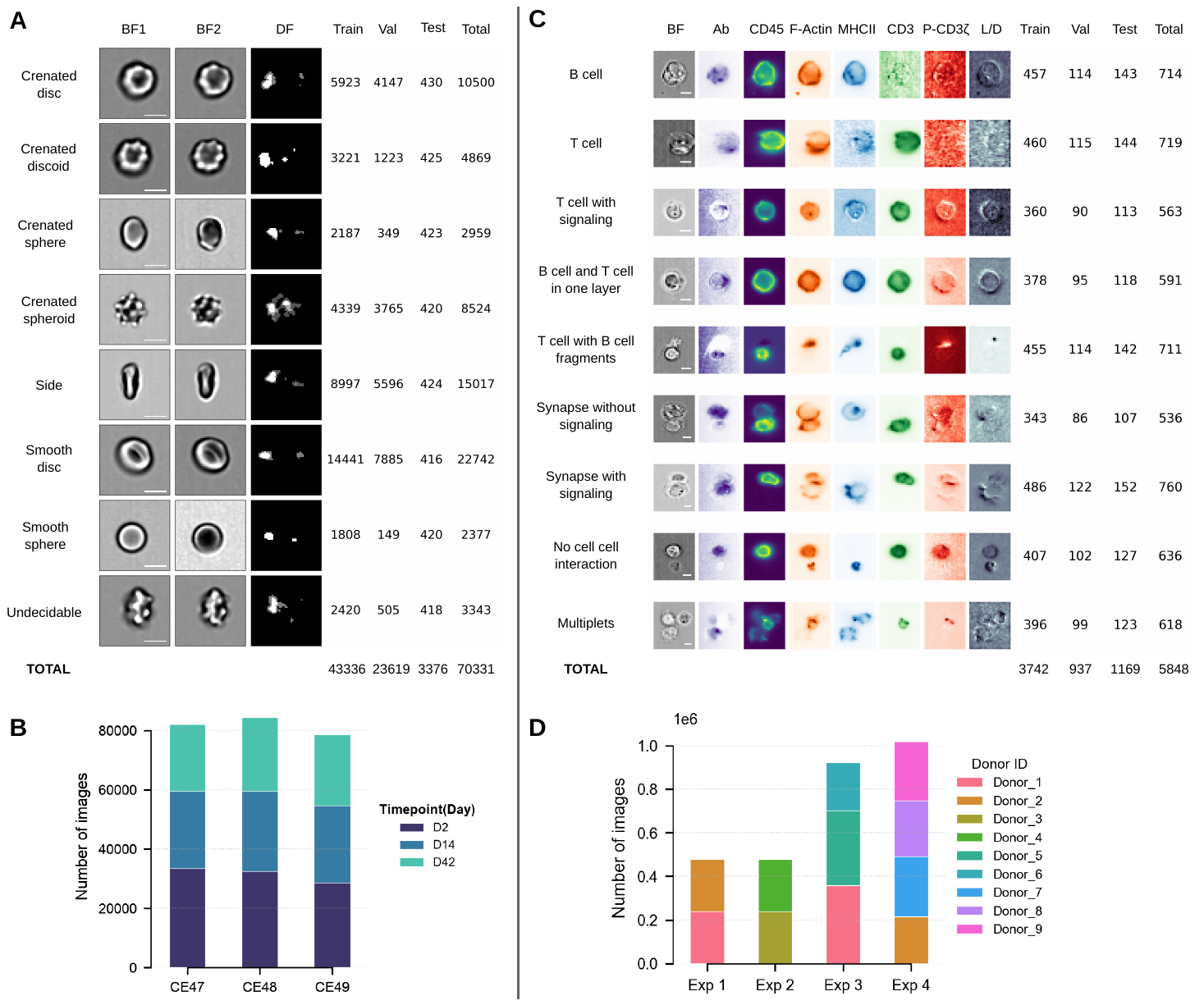}}
\caption{Overview of \gls{ifc} datasets and distribution of annotated and unannotated subsets. \textbf{(A)} Representative multichannel imagery and class distribution of the annotated \acrshort{rbc} dataset~\citep{doan_objective_2020}. For each class, three imaging channels are shown alongside the sample counts for training, validation, and testing splits. \textbf{(B)} Distribution of the \acrshort{rbc} unannotated subset across acquisition batches (bags (CE) and timepoints (Day)). Stacked bars represent the sample volume per CE, partitioned by day. \textbf{(C)} Representative multichannel imagery and class distribution of the annotated \acrshort{isf} dataset~\citep{essig_imaging_2022, shetab_boushehri_explainable_2023}. Training, validation, and testing counts are indicated per class. \textbf{(D)} Distribution of the \acrshort{isf} unannotated subset across four independent experiments. Stacked bars illustrate donor diversity within each experiment, showing the contribution of individual donors to the total unannotated pool. Scale bars in \textbf{(A)} and \textbf{(C)} represent $5\mu m$. \label{fig:experiments_datasets}}
\end{figure*}

For our benchmarking, we selected two publicly available \gls{ifc} datasets that represent distinct biological and computational challenges, ensuring the generalizability of our \gls{ssl} framework (Methods, Data Preparation). All images and annotations were obtained from the public releases of the original studies. First, the \gls{rbc} dataset~\citep{doan_objective_2020}, a specialized single-cell morphological dataset comprising $315,416$ multi-channel images ($\mathrm{BF1}$, $\mathrm{BF2}$, and $\mathrm{DF}$) from which $70,331$ were annotated by the original authors and categorized into eight distinct red blood cell shapes (\autoref{fig:experiments_datasets}A, B). Second, we leveraged the \gls{isf} dataset, described by its authors as one of the largest publicly available \gls{ifc} datasets, containing over $2.89$ million images profiling human immunological synapse formation~\citep{essig_imaging_2022, shetab_boushehri_explainable_2023} (\autoref{fig:experiments_datasets}D). A subset of $9,186$ images was manually annotated by the original authors  into $17$ classes. Similarly to \citep{shetab_boushehri_explainable_2023,boushehri2024pxpermute}, we focus on the nine biologically relevant classes (B cell, T cell, T cell with signaling, B cell and T cell in one layer, T cell with B cell fragments, Synapse without signaling, Synapse with signaling, No cell cell interaction, and Multiplets) as well as on the five canonical channels $\mathrm{BF}$, $\mathrm{F\mbox{--}Actin}$, $\mathrm{MHCII}$, $\mathrm{CD3}$, and $\mathrm{P\mbox{--}CD3\zeta}$ to ensure comparability with established baselines  (\autoref{fig:experiments_datasets}C). We refer to this specific five-channel configuration as \gls{isf}\mbox{--}5CH throughout the manuscript. Together, these datasets establish a rigorous cross-task performance landscape, spanning from single-cell morphology to complex intercellular signaling, and allowing for a comprehensive assessment of \gls{ssl} representation quality. We benchmarked six state-of-the-art \gls{ssl} methods (SimCLR~\citep{DBLP:conf/icml/ChenK0H20}, MoCo-v2~\citep{DBLP:journals/corr/abs-2003-04297}, MoCo-v3~\citep{DBLP:conf/iccv/ChenXH21}, DINO~\citep{DBLP:conf/iccv/CaronTMJMBJ21}, \gls{bt}~\citep{DBLP:conf/icml/ZbontarJMLD21}, and \gls{byol}~\citep{DBLP:conf/nips/GrillSATRBDPGAP20}) across four foundational architectures (ResNet50~\citep{DBLP:conf/cvpr/HeZRS16}, \gls{vit}\mbox{--}S/16~\citep{DBLP:conf/iclr/DosovitskiyB0WZ21}, ConvNeXtV1\mbox{--}T~\citep{DBLP:conf/cvpr/0003MWFDX22} and ConvNeXtV2\mbox{--}T~\citep{DBLP:conf/cvpr/WooDHC0KX23}), which can be initialized either with random weights ($\theta_{\mathrm{Rnd}}$) or ImageNet-supervised pretrained weights ($\theta_{\mathrm{IN}}$). These \gls{ssl} methods (Supplementary \gls{ssl}) represent three established families of self-supervision (contrastive, asymmetric student-teacher, and redundancy reduction) identified by recent studies for achieving state-of-the-art representation quality~\citep{balestriero2023cookbook}.

We simulate the acute annotation scarcity typical of specialized bioimaging domains, where high-fidelity manual annotations are labor-intensive and scarce, by evaluating each model across a graded sequence of annotated data fractions ($1\%, 5\%, 10\%, 25\%, 50\%$, and $100\%$). To compare downstream adaptation capacity while conserving computational resources, we restricted full-parameter finetuning to the best configurations identified for each weight initialization ($\mathrm{\theta_{Rnd}}$ or $\mathrm{\theta_{IN}}$) and evaluation protocol (\gls{knn}, linear probing, and XGBoost). The best configurations are systematically selected throughout the manuscript based on achieved macro F1-score under the fully annotated dataset regime ($100\%$ of $\mathcal{D}_{\mathrm{ann}}$).

\subsection*{Supervised Baseline Performance}\label{subsec:results_supervised_baselines}
\begin{figure*}[!htb]
\centerline{\includegraphics[width=\linewidth]{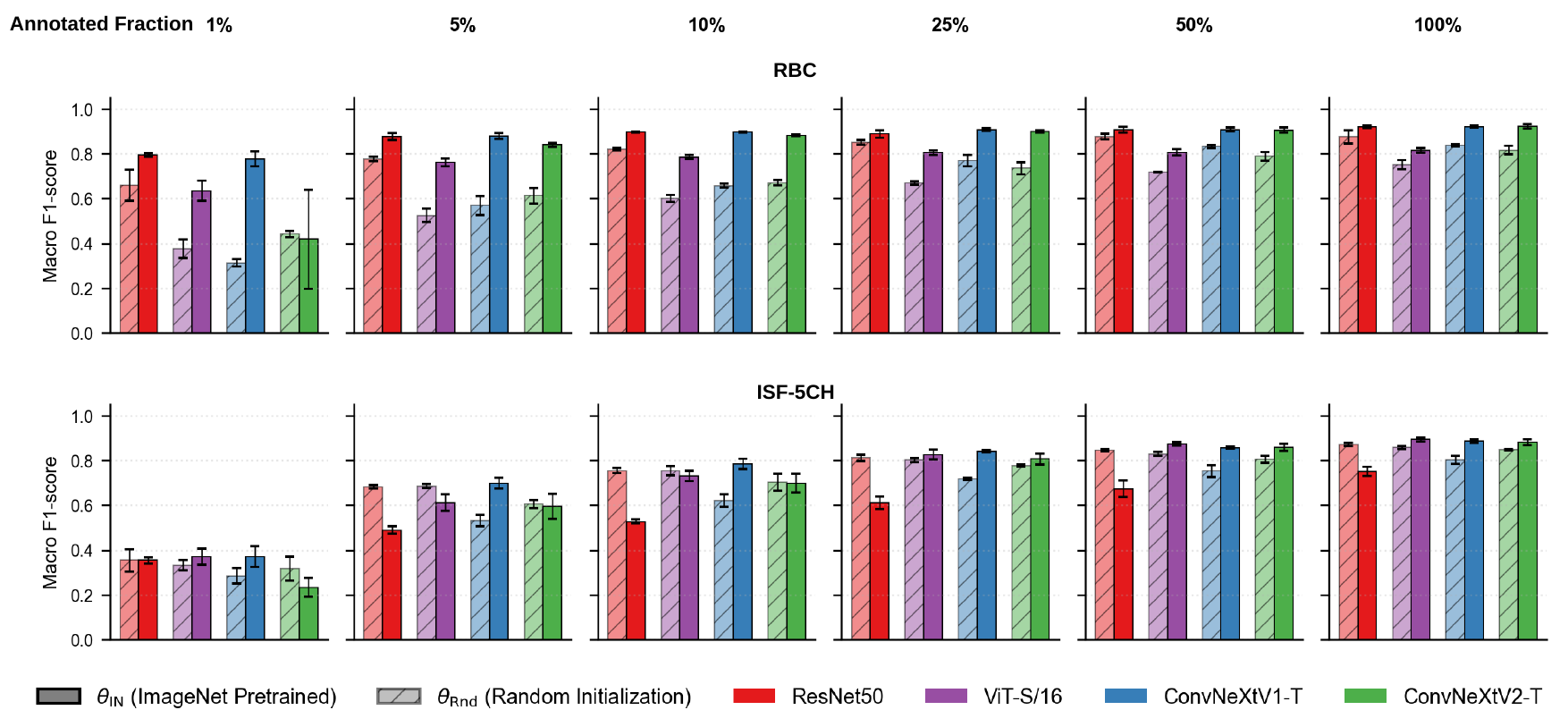}}
\caption{Test set performance of supervised baselines for both datasets (\gls{rbc} and \gls{isf}\mbox{--}5CH) across backbone variants, annotated data fractions, and weights initialization strategies\label{fig:supervised_baselines}}
\end{figure*}

To establish baselines, supervised models were trained end-to-end for each backbone architecture under the two initialization regimes $\theta_{\mathrm{Rnd}}$ and $\theta_{\mathrm{IN}}$. To quantify the baseline sensitivity to annotation scarcity, each baseline was systematically trained across diminishing annotated fractions of the training subset of $\mathcal{D}_{\mathrm{ann}}$, ranging from the full dataset to the low-data regime ($1\%$), and evaluated on the full test subset of $\mathcal{D}_{\mathrm{ann}}$. As shown in \autoref{tab:res:supervised:baelines} and further visualized in \autoref{fig:supervised_baselines},  \autoref{fig:supp_supp_baselines}A, and \autoref{fig:supp_supp_baselines}B, performance patterns varied across architectures, initialization, and data regimes. Across both datasets, all investigated backbones exhibited a highly consistent trajectory: As the fraction of annotated training data increased, test set macro F1-scores rose systematically while the performance gap between $\theta_{\mathrm{Rnd}}$ and $\theta_{\mathrm{IN}}$ models generally decreased, with architecture dependent exceptions.

{\setlength{\tabcolsep}{2pt}   
\renewcommand{\arraystretch}{0.95} 
\begin{table}[!htb]
\centering
\caption{Test set macro F1-scores of the supervised end-to-end baselines (mean $\pm$ std) on $100\%$ annotated data fraction. $\theta_{\mathrm{IN}}$ and $\theta_{\mathrm{Rnd}}$ denote initialization from ImageNet-supervised weights and random initialization, respectively. The best-performing supervised baseline for each dataset is \textbf{bolded}.
\label{tab:res:supervised:baelines}}
\begin{center}
\begin{tabular*}{240pt}{@{\extracolsep\fill}lcc@{\extracolsep\fill}}
\toprule
\textbf{Backbone} & \textbf{\gls{rbc}} & \textbf{\gls{isf}\mbox{--}5CH} \\
\midrule
ResNet50\mbox{--}$\theta_{\mathrm{Rnd}}$ & $0.875 \pm 0.030$ & $0.870 \pm 0.007$ \\
ResNet50\mbox{--}$\theta_{\mathrm{IN}}$& $0.920 \pm 0.006$ & $0.752 \pm 0.021$ \\
\midrule
ConvNeXtV1\mbox{--}T\mbox{--}$\theta_{\mathrm{Rnd}}$ & $0.839 \pm 0.006$ & $0.802 \pm 0.018$ \\
ConvNeXtV1\mbox{--}T\mbox{--}$\theta_{\mathrm{IN}}$ & $0.921 \pm 0.005$ & $0.889 \pm 0.007$ \\
ConvNeXtV2\mbox{--}T\mbox{--}$\theta_{\mathrm{Rnd}}$& $0.816 \pm 0.019$ & $0.848 \pm 0.002$ \\
ConvNeXtV2\mbox{--}T\mbox{--}$\theta_{\mathrm{IN}}$& $\mathbf{0.922 \pm 0.011}$ & $0.877 \pm 0.011$ \\
\midrule
\gls{vit}\mbox{--}S/16\mbox{--}$\theta_{\mathrm{Rnd}}$ & $0.752 \pm 0.020$ & $0.857 \pm 0.005$ \\
\gls{vit}\mbox{--}S/16\mbox{--}$\theta_{\mathrm{IN}}$ & $0.816 \pm 0.011$ & $\mathbf{0.893 \pm 0.010}$ \\

\bottomrule
\end{tabular*}
\end{center}
\end{table}
}

On the \gls{rbc} dataset, the impact of initialization revealed a clear pattern: Across all evaluated backbones, $\theta_{\mathrm{IN}}$ initialization consistently yielded performance improvements, with the singular exception of ConvNeXtV2\mbox{--}T at the sparse $1\%$ annotated data fraction (\autoref{fig:supervised_baselines}, \autoref{fig:supp_supp_baselines}A). The magnitude of this initialization advantage varied largely across backbone families. For ResNet50, the performance gap between $\theta_{\mathrm{IN}}$ and $\theta_{\mathrm{Rnd}}$ was the most constrained across all fractions, reaching its maximum delta in F1-score of $13.5$ \gls{pp} at the $1\%$ annotated data regime. Conversely, the most pronounced initialization sensitivity was observed within the ConvNeXt family. Under the $1\%$ data regime, ConvNeXtV1\mbox{--}T exhibited a $46.3$ \gls{pp} performance advantage under $\theta_{\mathrm{IN}}$ compared to $\theta_{\mathrm{Rnd}}$, with this margin narrowing to $31.0$, $24.0$, and $13.9$ at the $5\%$, $10\%$, and $25\%$ annotated data fractions, respectively. While ConvNeXtV2\mbox{--}T maintained a performance advantage of at least $10.0$ \gls{pp} under $\theta_{\mathrm{IN}}$ across higher fractions, $\theta_{\mathrm{Rnd}}$ slightly out−performed $\theta_{\mathrm{IN}}$ at the $1\%$ annotated scarcity level by $2.3$ \gls{pp}.  The \gls{vit}\mbox{--}S/16 demonstrated a monotonic decrease in initialization sensitivity as the volume of annotated training data scaled, with the $\theta_{\mathrm{IN}}$ advantage steadily declining from $25.9$ \gls{pp} at the $1\%$ annotated fraction to $6.4$ \gls{pp} under full $\mathcal{D}_{\mathrm{ann}}$.

On the \gls{isf}\mbox{--}5CH dataset, the impact of initialization strategies revealed a more heterogeneous distribution of performance across the backbones and data regimes. ResNet50 consistently benefited from $\theta_{\mathrm{Rnd}}$ initialization, except for the $1\%$ annotated data fraction where no difference was observed (\autoref{fig:supervised_baselines}). The largest delta in F1-score is observed at $10\%$ of $\mathcal{D}_{\mathrm{ann}}$ with $22.8$ \gls{pp} in favor of $\theta_{\mathrm{Rnd}}$ (\autoref{fig:supp_supp_baselines}A). Conversely, ConvNeXtV1\mbox{--}T demonstrated a consistent preference for $\theta_{\mathrm{IN}}$ across all fractions with a minimal delta of $8.7$ \gls{pp}. Both ConvNeXtV2\mbox{--}T and \gls{vit}\mbox{--}S/16 exhibited a transitional behavior: While performance fluctuated between initialization strategies in low-data regimes ($1\mbox{--}10\%$ of $\mathcal{D}_{\mathrm{ann}}$), $\theta_{\mathrm{IN}}$ pretraining provided a minimal performance advantage once the annotated fraction reached $25\%$ or higher (\autoref{fig:supervised_baselines}, \autoref{fig:supp_supp_baselines}A). 

To investigate whether integrating domain-specific knowledge enhances feature extraction, we applied during training the domain-specific preprocessing Logicle transformation~\citep{parks_new_2006, moore_update_2012} following the Gating-ML 2.0 standard~\citep{https://doi.org/10.1002/cyto.a.22690} to the supervised baseline. Contrary to our initial hypothesis, the inclusion of the Logicle transform during training did not yield any performance gains (\autoref{fig:supp_supp_baselines}C). Instead, we observed a slight decrease in F1-score, demonstrating that, while non-linear, domain-specific transformations remain a gold standard for human-in-the-loop visualization (\autoref{fig:method} Data preparation) and manual gating~\citep{moore_update_2012}, they are detrimental to deep representational learning.

Interestingly, we observed on the \gls{rbc} dataset that the supervised baseline already achieved solid performance when trained on just $1\%$ of $\mathcal{D}_{\mathrm{ann}}$. To identify the breakdown threshold of standard supervised learning and verify if \gls{ssl} provides a margin of benefit under extreme constraints, we established a $0.1\%$ ultra-low annotated fraction of $\mathcal{D}_{\mathrm{ann}}$ comparison (\autoref{fig:supp_supp_baselines}B). Similarly to the \gls{isf}\mbox{--}5CH dataset at this scarcity level, the best supervised baseline (ResNet50\mbox{--}$\theta_{\mathrm{IN}}$) achieves a very low F1-score of $43.7 \pm 1.0 \%$. Furthermore, the vulnerability in the newer ConvNeXtV2\mbox{--}T architecture under extreme data scarcity is pronounced, with the performance advantage of $\theta_{\mathrm{Rnd}}$ against $\theta_{\mathrm{IN}}$, doubling from $2.3$ \gls{pp} $\Delta F1$ at the $1\%$ annotated scarcity level to $4.6$ \gls{pp} $\Delta F1$ at $0.1\%$ annotated scarcity. This choice of fractions balances the absolute sample size between the two datasets, yielding $37$ training samples at the $1\%$ fraction for the \gls{isf}\mbox{--}5CH dataset and $43$ samples at the $0.1\%$ fraction for the \gls{rbc} dataset (as low as two samples per class), thereby enabling a fairer and more balanced comparison of representation capacity under equivalent low-data regimes. Having established the performance bounds and architectural sensitivities of the supervised baselines, we next evaluated whether self-supervised pretraining can bridge the observed domain gap and stabilize representation learning across modalities.

\subsection*{Self-Supervised Pretraining from Scratch Bridges the Domain Gap for Immunological Synapses}\label{subsec:results_isf_pt}
\begin{figure*}[!htb]
\centerline{\includegraphics[width=\linewidth]{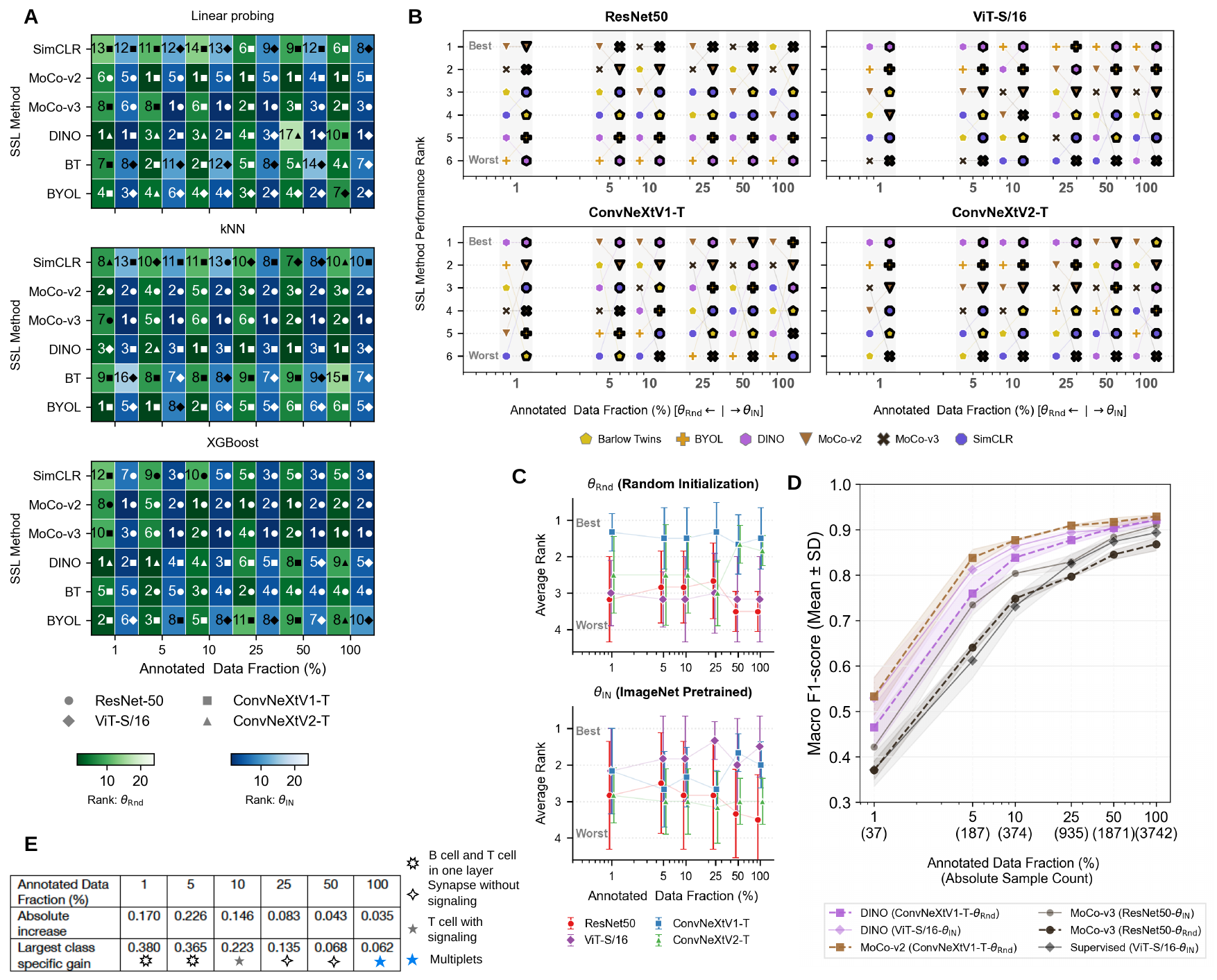}}
\caption{ 
Benchmarking self-supervised representations for the \gls{isf}\mbox{--}5CH dataset. \textbf{(A)} Ranking of combinations of \gls{ssl} methods and backbones across downstream evaluation protocols. Heatmaps display the rank of $48$ model combinations (six \gls{ssl} methods times four backbones times two initialization strategies) across \gls{knn}, linear probing, and XGBoost protocols for each annotated fraction of $\mathcal{D}_{\mathrm{ann}}$. Each cell is split to compare initialization $\theta_{\mathrm{Rnd}}$ (left/green, $24$ ranks) vs. $\theta_{\mathrm{IN}}$ (right/blue, $24$ ranks). Darker hues indicate superior ranking (rank 1). Markers identify the corresponding backbone architecture for the specific configuration. \textbf{(B)} \gls{ssl} method performance ranking for each backbone from each annotated fraction based on linear probing evaluation. Markers with bolded edges represent $\theta_{\mathrm{IN}}$ and normal edges $\theta_{\mathrm{Rnd}}$. \textbf{(C)} Architectural robustness and initialization bias: Comparison of average backbone rank after linear probing between initializations $\theta_{\mathrm{Rnd}}$ and $\theta_{\mathrm{IN}}$. Error bars indicate the inter \gls{ssl} method standard deviation. \textbf{(D)} Finetuning trajectories of top-ranked configurations. For each initialization mode ($\theta_{\mathrm{Rnd}}$ or $\theta_{\mathrm{IN}}$) and evaluation protocol, the best-performing \gls{ssl}-backbone pairing as indicated in \autoref{tab:res:ssl:ISF-5CH_DS:f1_knn_lp_xgb} was selected. \textbf{(E)} Finetuning performance gain between the best supervised baseline at $100\%$ of $\mathcal{D}_{\mathrm{ann}}$ (\gls{vit}\mbox{--}S/16\mbox{--}$\theta_{\mathrm{IN}}$) and the best \gls{ssl} combination (ConvNeXtV1\mbox{--}T\mbox{--}$\theta_{\mathrm{Rnd}}$ with MoCo-v2). The table displays the absolute improvements (\gls{pp} delta) in overall macro-F1-score along with the single class showing the most substantial improvement for each annotated fraction of $\mathcal{D}_{\mathrm{ann}}$. 
Fractions on the x-axis in panels \textbf{(B)}, \textbf{(C)}, and \textbf{(D)} are displayed on a logarithmic scale for clarity in low-data regimes. Ranks (1 = top performer) per annotated fraction ($1\% \mbox{--} 100\%$) in \textbf{(B)} and \textbf{(C)} were calculated based on test set macro F1-scores using a linear probing protocol.\label{fig:results_ranking_isf}}
\end{figure*}

Across all three major evaluation protocols (Methods, Downstream Analysis) (\gls{knn}, linear probing, and XGBoost), we noticed a consistent performance advantage for backbones trained via self-supervision from $\theta_{\mathrm{Rnd}}$ compared to those initialized with $\theta_{\mathrm{IN}}$. Specifically, the target-domain \gls{ssl} approach yielded median improvements in F1-score on the full $\mathcal{D}_{\mathrm{ann}}$ of $2.3$ \gls{pp} for \gls{knn}, $2.0$ \gls{pp} for linear probing, and $4.2$ \gls{pp} for XGBoost (\autoref{tab:res:ssl:ISF-5CH_DS:f1_knn_lp_xgb}). Observing the general methodological performance across all evaluated backbones, DINO, MoCo-v2, and MoCo-v3 emerge with the highest performance ranks across several annotation scarcity regimes (\autoref{fig:results_ranking_isf}A, B, \autoref{tab:res:ssl:ISF-5CH_DS:f1_knn_lp_xgb}), with the best \gls{ssl} configuration  (ConvNeXtV1\mbox{--}T\mbox{--}$\theta_{\mathrm{Rnd}}$ with MoCo-v2) being only $1.4$ \gls{pp} below the best supervised baseline in F1-score when evaluated by linear probing. The second best (\gls{vit}\mbox{--}S/16\mbox{--}$\theta_{\mathrm{IN}}$ with DINO) was also not far from the best supervised baseline ($2.1$ \gls{pp}). These narrow gaps provide strong quantitative evidence of the linear separability of the learned feature space. Most notably, DINO achieved on all three evaluation protocols the top ranking across the \gls{vit}\mbox{--}S/16, ConvNeXtV1\mbox{--}T, and ConvNeXtV2\mbox{--}T backbones at the $1\%$ annotated data fraction, holding true under both $\theta_{\mathrm{Rnd}}$ and $\theta_{\mathrm{IN}}$ (\autoref{fig:results_ranking_isf}B,  \autoref{fig:supp_s3_isf_knn}, \autoref{fig:supp_s3_isf_lin_probe},  \autoref{fig:supp_s3_isf_xgboost}). The architectural choice also plays a critical role in how cleanly these self-supervised representations align with the downstream task. The average model rank when pretraining from scratch ($\theta_{\mathrm{Rnd}}$) is led by ConvNeXtV1\mbox{--}T, which displays the most stable and robust representations under strict data limits, whereas \gls{vit}\mbox{--}S/16 dominates other backbones under $\theta_{\mathrm{IN}}$ (\autoref{fig:results_ranking_isf}C). ResNet50 backbone exhibits the poorest performance and stability. Specifically, it ranks third overall under $\theta_{\mathrm{Rnd}}$ and becomes last when transferred from $\theta_{\mathrm{IN}}$, accompanied by wide standard deviation bars, indicating high susceptibility to initialization variance. This volatility is attributed to its high sensitivity to the choice of self-supervised optimization objective (\autoref{tab:res:ssl:ISF-5CH_DS:f1_knn_lp_xgb}, \autoref{fig:results_ranking_isf}B). While ResNet50 exhibits strong representations when paired with contrastive methods (SimCLR, MoCo-v2, MoCo-v3), it demonstrates susceptibility to representation collapse or dimensional degradation when subjected to asymmetric or redundancy-reduction (DINO, \gls{bt}, \gls{byol}) (\autoref{tab:res:ssl:ISF-5CH_DS:f1_knn_lp_xgb}, \autoref{fig:supp_s3_isf_knn}, \autoref{fig:supp_s3_isf_lin_probe}, \autoref{fig:supp_s3_isf_xgboost}), with some or all of the three runs not converging.

Zooming into specific annotated fractions, these patterns become more pronounced. A closer inspection of the low-data regime reveals that self-supervised pretraining closes the gap to, and in some cases surpasses, traditional supervised learning under extreme annotation scarcity (\autoref{fig:supp_s3_isf_knn}, \autoref{fig:supp_s3_isf_lin_probe} \autoref{fig:supp_s3_isf_xgboost}). Specifically, at $1\%$ annotated fraction of $\mathcal{D}_{\mathrm{ann}}$, pretraining with DINO or MoCo-v2 combined with downstream linear probing or XGBoost classification outperforms the standard end-to-end supervised baseline. Under $\theta_{\mathrm{IN}}$, the best supervised baseline (\gls{vit}\mbox{--}S/16 with $37.1 \pm 3.6\%$ F1-score) was outperformed by $1.7$ \gls{pp} when pretraining a ResNet50 with MoCo-v2 followed by XGBoost, and by $8.9$ \gls{pp} when combining DINO with ConvNeXtV1\mbox{--}T and applying linear probing. This margin is amplified under $\theta_{\mathrm{Rnd}}$ where DINO with ConvNeXtV2\mbox{--}T achieved $40.8 \pm 3.1\%$ after XGBoost and $47.4\% \pm 6.5\%$ after linear probing, both exceeding the best supervised baseline (ResNet50 with $35.4 \pm 5.1\%$), demonstrating that the learned representations capture highly discriminative signal features even when downstream annotation availability is minimal. From $5\%$ annotated $\mathcal{D}_{\mathrm{ann}}$ onward, ConvNeXtV1\mbox{--}T\mbox{--}$\theta_{\mathrm{Rnd}}$ with MoCo-v2 pretraining followed by linear probing outperforms the best end-to-end supervised baseline under $\theta_{\mathrm{Rnd}}$ (\autoref{fig:supp_s3_isf_lin_probe}). This shows that self-supervised pretraining approaches the performance of a supervised baseline or a fully finetuned \gls{ssl} model without the computational burden of end-to-end training. While frozen latent space evaluations are not expected to exceed the upper bound of full \gls{ssl} finetuning, they remain highly competitive even under extreme annotation scarcity, a regime where supervised baselines fail to generalize.

{\setlength{\tabcolsep}{1pt}   
\renewcommand{\arraystretch}{0.6} 
\begin{table*}[!t]
\centering

\caption{Evaluation of the \gls{ssl} methods on the fully annotated \gls{isf}\mbox{--}5CH dataset: Test set macro F1-scores (mean $\pm$ std) for \gls{knn} evaluation, linear probing, and XGBoost. $\theta_{\mathrm{IN}}$ and $\theta_{\mathrm{Rnd}}$ denote initialization from ImageNet-supervised weights and random initialization, respectively. For each evaluation protocol and weight initialization, the best score per backbone is \underline{underlined}, and the best score across all backbones is highlighted in \textbf{bold}.\label{tab:res:ssl:ISF-5CH_DS:f1_knn_lp_xgb}}
\begin{threeparttable}
\begin{tabular*}{\textwidth}{@{\extracolsep\fill} l l c c c c c c @{\extracolsep\fill}}
\toprule
\textbf{Backbone} & \textbf{Method}
  & \multicolumn{2}{c}{\textbf{\gls{knn}}}
  & \multicolumn{2}{c}{\textbf{Linear probing}}
  & \multicolumn{2}{c}{\textbf{XGBoost}} \\
\cmidrule(lr){3-4} \cmidrule(lr){5-6} \cmidrule(lr){7-8}
&
  & $\mathbf{\theta_{Rnd}}$ & $\mathbf{\theta_{IN}}$
  & $\mathbf{\theta_{Rnd}}$ & $\mathbf{\theta_{IN}}$
  & $\mathbf{\theta_{Rnd}}$ & $\mathbf{\theta_{IN}}$ \\
\midrule
\multirow{7}{*}{ResNet50}
  & SimCLR  & $0.633\pm0.011$ & $0.639\pm0.007$ & $0.845\pm0.020$ & $0.799\pm0.006$ & $0.832\pm0.005$ & $0.824\pm0.005$ \\
  & MoCo-v2  & $0.710\pm0.012$ & $0.740\pm0.013$ & $0.844\pm0.015$ & $0.831\pm0.005$ & $0.849\pm0.006$ & $0.859\pm0.012$ \\
  & MoCo-v3  & $\underline{0.722\pm0.006}$ & $\underline{\mathbf{0.760\pm0.018}}$ & $0.852\pm0.009$ & $\underline{0.856\pm0.014}$ & $\underline{\mathbf{0.850\pm0.010}}$ & $\underline{\mathbf{0.867\pm0.012}}$ \\
  & DINO    & $0.495\pm0.018$ & $0.299\pm0.017$ & $0.717\pm0.005$ & $0.433\pm0.023$ & $0.688\pm0.010$ & $0.449\pm0.006$ \\
  & \gls{bt}      & $0.662\pm0.013$ & $0.599\pm0.003$ & $\underline{0.857\pm0.010}$ & $0.807\pm0.013$ & $0.840\pm0.010$ & $0.820\pm0.001$ \\
  & \gls{byol}    & $0.290\pm0.007$ & $0.599\pm0.008$ & $0.448\pm0.010$ & $0.660\pm0.009$ & $0.532\pm0.011$ & $0.751\pm0.004$ \\
\midrule
\multirow{7}{*}{ConvNeXtV1\mbox{--}T}
  & SimCLR  & $0.668\pm0.009$ & $0.647\pm0.004$ & $0.862\pm0.012$ & $0.824\pm0.005$ & $0.789\pm0.003$ & $0.760\pm0.005$ \\
  & MoCo-v2  & $0.687\pm0.014$ & $0.657\pm0.004$ & $\underline{\mathbf{0.879\pm0.010}}$ & $\underline{0.850\pm0.005}$ & $\underline{0.844\pm0.007}$ & $\underline{0.795\pm0.007}$ \\
  & MoCo-v3  & $0.677\pm0.006$ & $0.637\pm0.003$ & $0.870\pm0.003$ & $0.828\pm0.002$ & $0.827\pm0.007$ & $0.759\pm0.015$ \\
  & DINO    & $\underline{\mathbf{0.723\pm0.012}}$ & $\underline{0.691\pm0.005}$ & $0.854\pm0.009$ & $0.837\pm0.010$ & $0.806\pm0.004$ & $0.785\pm0.010$\\
  & \gls{bt}      & $0.665\pm0.003$ & $0.645\pm0.011$ & $0.857\pm0.014$ & $0.832\pm0.011$ & $0.805\pm0.009$ & $0.757\pm0.012$ \\
  & \gls{byol}    & $0.691\pm0.012$ & $0.666\pm0.009$ & $0.817\pm0.005$ & $0.851\pm0.009$ & $0.805\pm0.007$ & $0.757\pm0.013$ \\
\midrule
\multirow{7}{*}{ConvNeXtV2\mbox{--}T}
  & SimCLR  & $0.676\pm0.013$ & $0.634\pm0.003$ & $0.852\pm0.008$ & $0.822\pm0.005$ & $0.789\pm0.003$ & $0.744\pm0.001$ \\
  & MoCo-v2  & $0.665\pm0.004$ & $0.640\pm0.008$ & $\underline{0.867\pm0.008}$ & $0.836\pm0.009$ & $\underline{0.820\pm0.009}$ & $0.780\pm0.004$ \\
  & MoCo-v3  & $0.676\pm0.011$ & $0.625\pm0.009$ & $0.863\pm0.008$ & $0.800\pm0.009$ & $0.803\pm0.010$ & $0.725\pm0.013$ \\
  & DINO    & $\underline{0.703\pm0.008}$ & $\underline{0.661\pm0.002}$ & $0.846\pm0.006$ & $0.826\pm0.009$ & $0.812\pm0.001$ & $\underline{0.790\pm0.012}$ \\
  & \gls{bt}      & $0.661\pm0.003$ & $0.635\pm0.010$ & $0.863\pm0.005$ & $\underline{0.837\pm0.004}$ & $0.774\pm0.008$ & $0.752\pm0.009$ \\
  & \gls{byol}    & $0.654\pm0.019$ & $0.599\pm0.016$ & $0.851\pm0.008$ & $0.824\pm0.005$ & $0.813\pm0.009$ & $0.716\pm0.018$ \\
\midrule
\multirow{7}{*}{\gls{vit}\mbox{--}S/16}
  & SimCLR  & $0.668\pm0.007$ & $0.646\pm0.015$ & $0.828\pm0.005$ & $0.839\pm0.010$ & $0.770\pm0.004$ & $0.757\pm0.010$ \\
  & MoCo-v2  & $0.672\pm0.009$ & $0.638\pm0.006$ & $0.852\pm0.010$ & $0.845\pm0.011$ & $0.786\pm0.005$ & $0.764\pm0.007$ \\
  & MoCo-v3  & $0.658\pm0.006$ & $0.644\pm0.011$ & $0.846\pm0.012$ & $0.816\pm0.024$ & $0.784\pm0.008$ & $0.759\pm0.004$ \\
  & DINO    & $\underline{0.692\pm0.013}$ & $\underline{0.709\pm0.017}$ & $0.823\pm0.018$ & $\underline{\mathbf{0.872\pm0.003}}$ & $0.786\pm0.011$ & $\underline{0.809\pm0.009}$ \\
  & \gls{bt}      & $0.649\pm0.006$ & $0.662\pm0.004$ & $0.843\pm0.012$ & $0.840\pm0.007$ & $0.775\pm0.005$ & $0.770\pm0.002$ \\
  & \gls{byol}    & $0.655\pm0.017$ & $0.671\pm0.007$ & $\underline{0.859\pm0.000}$ & $0.860\pm0.009$ & $\underline{0.801\pm0.010}$ & $0.777\pm0.018$ \\
\midrule
\multicolumn{2}{l}{\textbf{Overall median}}
  & $0.668$ & $0.645$& $0.852$ & $0.832$ & $0.804$& $0.762$ \\
\bottomrule
\end{tabular*}
\end{threeparttable}
\end{table*}
}

\subsection*{Self-Supervised Learning Resolves Complex Immunological Structures in Sparse Annotation Regimes}\label{subsec:results_isf_ft}
A persistent challenge in \gls{ifc} is the understanding of complex structures, particularly the accurate detection of multi-cellular events~\citep{https://doi.org/10.1002/cyto.a.24961}. Our results demonstrate that \gls{ssl} mitigates this bottleneck. Following our systematic evaluation of various \gls{ssl} configurations, we finetuned the leading ones to assess the transferability of their self-supervised features and to quantify their effectiveness as weight initializations for downstream classification. As shown in \autoref{fig:results_ranking_isf}D, finetuning our pretrained models yields improvements over the fully supervised baselines. The best combinations proposed by the XGBoost protocol (ResNet50\mbox{--}$\theta_{\mathrm{Rnd}}$ and ResNet50\mbox{--}$\theta_{\mathrm{IN}}$ both with MoCo-v3) showed the smallest improvements, with ResNet50\mbox{--}$\theta_{\mathrm{Rnd}}$ + MoCo-v3 even falling under the supervised baseline from $25\%$ annotated $\mathcal{D}_{\mathrm{ann}}$ onward (\autoref{fig:results_ranking_isf}D). The best finetuning performance was achieved with ConvNeXtV1\mbox{--}T\mbox{--}$\theta_{\mathrm{Rnd}}$ + MoCo-v2, surpassing the supervised baseline on full $\mathcal{D}_{\mathrm{ann}}$ already at $25\%$ (a little less than $1,000$ samples) of $\mathcal{D}_{\mathrm{ann}}$ ($90.8 \pm 0.1\%$ against $89.3\pm 1.0\%$ F1-score), and achieving $93.0 \pm0.3\%$ on full $\mathcal{D}_{\mathrm{ann}}$. This demonstrates that self-supervised pretraining serves as a robust initialization state that accelerates downstream convergence.

While these global metrics illustrate clear cross-fraction success, a granular, class-specific examination reveals where the true strength of self-supervised representations lies. When investigating the source of improvements between the supervised baseline and our top finetuned model (\autoref{fig:results_ranking_isf}D, E), we noticed that pretraining disentangles complex structures. At $1\%$ and $5\%$ of $\mathcal{D}_{\mathrm{ann}}$ where the largest improvements in F1-score occur, the largest class-specific gain stems from the class "B cell and T cell in one layer" (\autoref{fig:experiments_datasets}C) with deltas of $38.0$ \gls{pp} and $36.5$ \gls{pp} respectively (\autoref{fig:results_ranking_isf}E). This class corresponds to events in which the B and T cell completely overlap~\citep{shetab_boushehri_explainable_2023}; the resulting co‑localized high intensities create ambiguous feature profiles that commonly lead classifiers to assign these examples to single cell classes (in our case B cell or T cell with or without signaling)~\citep{cano2013introduction}. At $10\%$ and $25\%$ of $\mathcal{D}_{\mathrm{ann}}$, the boost comes from classes  "T cell with signaling" with $\Delta F1 = 22.3$ \gls{pp} and "Synapses without signaling" with $\Delta F1 = 13.5$ \gls{pp} respectively (\autoref{fig:results_ranking_isf}E), showing that \gls{ssl} separated signaling samples more cleanly. Biologically, these results suggest that self-supervised pretraining successfully learns to encode generalized spatial boundaries, cell-to-cell adhesion geometries, and relative morphological proportions from unannotated data. When the network is subsequently finetuned on scarce annotations, it possesses the necessary structural priors to identify complex cellular configurations.

\begin{figure*}[!htb]
\centerline{\includegraphics[width=\linewidth]{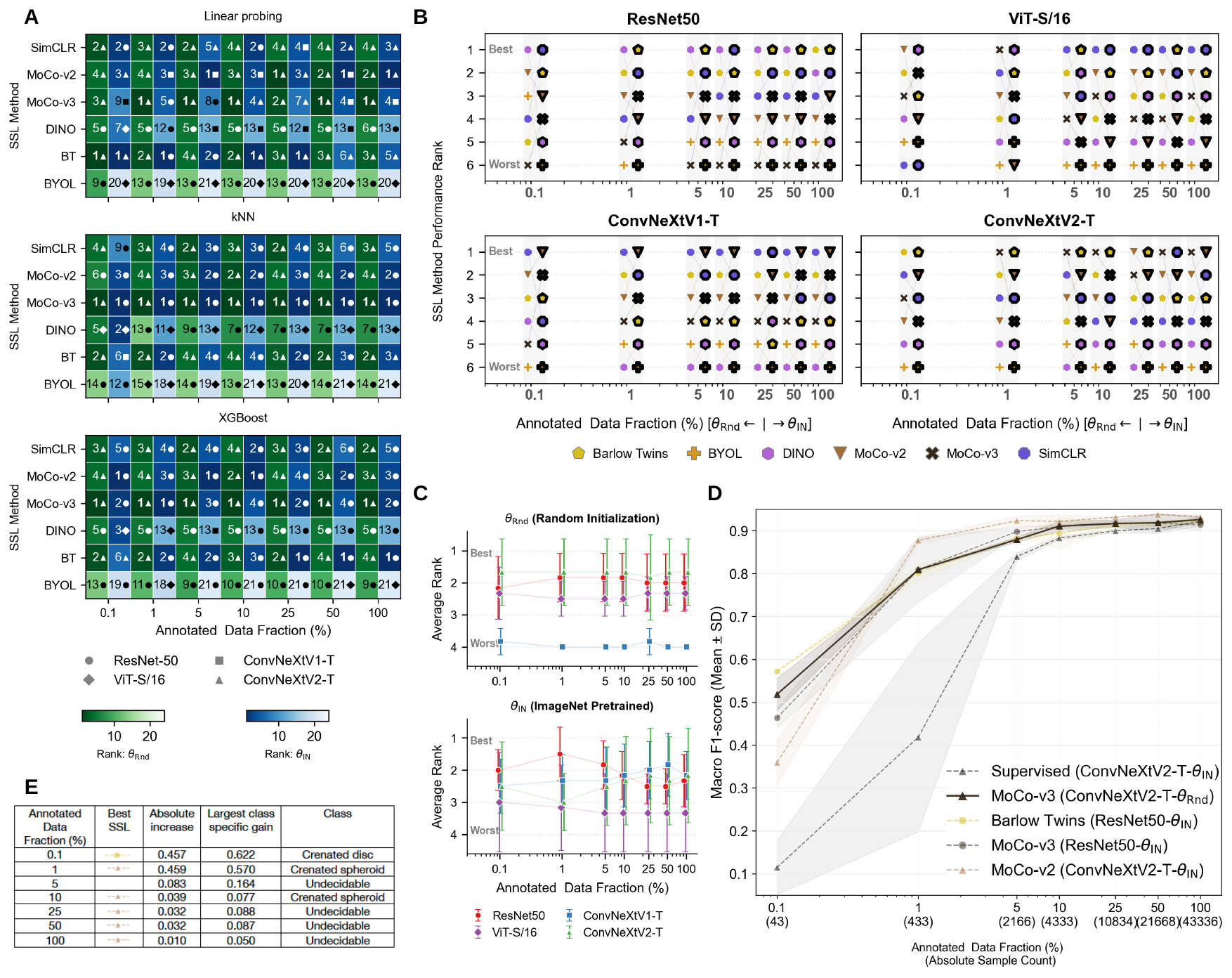}}
\caption{ 
Benchmarking self-supervised representations for the \gls{rbc} dataset. \textbf{(A)} Ranking of combinations of \gls{ssl} methods and backbones across downstream evaluation protocols. Heatmaps display the rank of $48$ model combinations (six \gls{ssl} methods times four backbones times two initialization strategies) across \gls{knn}, linear probing, and XGBoost protocols for each annotated fraction. Each cell is split to compare initialization $\theta_{\mathrm{Rnd}}$ (left/green, $24$ ranks) vs. $\theta_{\mathrm{IN}}$ (right/blue, $24$ ranks). Darker hues indicate superior ranking (rank 1). Markers identify the corresponding backbone architecture for the specific configuration. \textbf{(B)} \gls{ssl} method performance ranking for each backbone from each annotated fraction based on linear probing evaluation. Markers with bolded edges represent $\theta_{\mathrm{IN}}$ and normal edges $\theta_{\mathrm{Rnd}}$. \textbf{(C)} Architectural robustness and initialization bias: Comparison of average backbone rank after linear probing between initializations $\theta_{\mathrm{Rnd}}$ and $\theta_{\mathrm{IN}}$. Error bars indicate the inter \gls{ssl} method standard deviation. \textbf{(D)} Finetuning trajectories of top-ranked configurations. For each initialization mode ($\theta_{\mathrm{Rnd}}$ or $\theta_{\mathrm{IN}}$) and evaluation protocol, the best-performing \gls{ssl}-backbone pairing as indicated in \autoref{tab:res:ssl:RBC_DS:f1_knn_lp_ft_xgb_split} was selected. \textbf{(E)} Finetuning performance gain between the best supervised baseline at $100\%$ of $\mathcal{D}_{\mathrm{ann}}$ (ConvNeXtV2\mbox{--}T\mbox{--}$\theta_{\mathrm{IN}}$) and the best \gls{ssl} combination for each annotated fraction from \textbf{(E)}. The table displays the absolute improvements (percentage point delta) in overall macro-F1-score along with the single class showing the most substantial improvement for each annotated fraction of $\mathcal{D}_{\mathrm{ann}}$. 
Fractions on the x-axis in panels \textbf{(B)}, \textbf{(C)}, and \textbf{(D)} are displayed on a logarithmic scale for clarity in low-data regimes. Ranks (1 = top performer) per annotated fraction ($0.1\% \mbox{--} 100\%$) in \textbf{(B)} and \textbf{(C)} were calculated based on test set macro F1-scores using a linear probing protocol.\label{fig:results_ranking_rbc}}
\end{figure*}

\subsection*{ImageNet Pretraining Enhances Performance for Morphological Cell Shape Classification}\label{subsec:results_rbc}
On the \gls{rbc} dataset, conversely to the \gls{isf}\mbox{--}5CH dataset, initializing the backbones with $\theta_{\mathrm{IN}}$ yielded better performances, with median $\Delta F1$ for the full $\mathcal{D}_{\mathrm{ann}}$ of $24.8$ \gls{pp} on \gls{knn}, $22.5$ \gls{pp} on linear probing, and $22.5$ \gls{pp} on XGBoost (\autoref{tab:res:ssl:RBC_DS:f1_knn_lp_ft_xgb_split}). This indicates that ImageNet-pretrained initializations provide a substantial advantage to downstream \gls{ssl} for biological and clinical tasks where feature extraction heavily relies on morphological analysis. 

Further analysis revealed that contrastive methods (SimCLR, MoCo-v2 and MoCo-v3) and \gls{bt} perform better, particularly when using the ConvNeXtV2\mbox{--}T backbone initialized with $\theta_{\mathrm{IN}}$ (\autoref{fig:results_ranking_rbc}A). Similar to the \gls{isf}\mbox{--}5CH dataset, linear probing already achieves competitive performance, with the top \gls{ssl} combination (ConvNeXtV2\mbox{--}T\mbox{--}$\theta_{\mathrm{IN}}$ with MoCo-v2) being only $1.2$ \gls{pp} below the best supervised baseline in F1-score (ConvNeXtV2\mbox{--}T\mbox{--}$\theta_{\mathrm{IN}}$), followed by ConvNeXtV2\mbox{--}T\mbox{--}$\theta_{\mathrm{Rnd}}$ + MoCo-v3 with $1.7$ \gls{pp} (\autoref{tab:res:ssl:RBC_DS:f1_knn_lp_ft_xgb_split}). A visible trend is the general poor performance of \gls{byol} (\autoref{tab:res:ssl:RBC_DS:f1_knn_lp_ft_xgb_split}, \autoref{fig:results_ranking_rbc}A, B), confirming reported instability issues~\citep{pmlr-v139-tian21a}. We additionally observe that DINO performs poorly on the smaller \gls{rbc} dataset compared to its performance on the larger \gls{isf}\mbox{--}5CH dataset, unless it is paired explicitly with a \gls{vit} (\autoref{tab:res:ssl:ISF-5CH_DS:f1_knn_lp_xgb}, \autoref{tab:res:ssl:RBC_DS:f1_knn_lp_ft_xgb_split}). This pattern is consistent with DINO having been developed and tuned on \glspl{vit}, for which its multi-crop strategy and high-dimensional prototype head were introduced. The convolutional configurations share the objective but not the architecture it was designed around. Because method specific hyperparameters were mostly held at their published defaults, we cannot separate an intrinsic interaction between objective and architecture from a suboptimal configuration for the convolutional backbones.

Assessing the general backbone ranking (\autoref{fig:results_ranking_rbc}C) reveals the following: First, ConvNeXtV2\mbox{--}T emerged as the overall superior backbone, achieving the top average rank and reduced performance variance. This performance gap was particularly pronounced under $\theta_{\mathrm{Rnd}}$, where ConvNeXtV2\mbox{--}T consistently performed best across all evaluated data fractions. Second, ConvNeXtV1\mbox{--}T performed poorly under $\theta_{\mathrm{Rnd}}$, consistently ranking last across all annotated data fractions, yet benefited disproportionately from $\theta_{\mathrm{IN}}$ initialization, which effectively closed the performance gap and brought its ranking in line with other competitive backbones. Third, the \gls{vit}\mbox{--}S/16 exhibited poor overall performance, particularly under $\theta_{\mathrm{IN}}$ where it ranked last alongside exhibiting very high performance variance. Finally, ResNet50 demonstrated stable performance, maintaining a consistent overall average rank of two under both random and ImageNet initializations. Notably, under $\theta_{\mathrm{IN}}$, ResNet50 excelled as the top-performing model up to the $5\%$ annotated data fraction, after which its ranking slipped slightly to second place and even third place. This indicates that while classical residual architectures remain highly competitive and sample-efficient in low-data regimes, modern convolutional backbones like ConvNeXtV2\mbox{--}T ultimately scale better as more annotated data become available.

A closer look at the evaluation of the specific annotated fraction enables deriving some key insights. At ultra-low scarcity levels, applying \gls{knn} to ResNet50\mbox{--}$\theta_{\mathrm{IN}}$ + MoCo-v3 is only $1$ \gls{pp} below the best supervised baseline ResNet50\mbox{--}$\theta_{\mathrm{IN}}$ (\autoref{fig:supp_s3_rbc_knn}). Most notably, \gls{bt} achieves remarkable performances (\autoref{fig:supp_s3_rbc_lin_probe}). Under $\theta_{\mathrm{IN}}$ and with $0.1\%$ of $\mathcal{D}_{\mathrm{ann}}$, ConvNeXtV2\mbox{--}T + \gls{bt} followed by linear probing improves the best supervised baseline (ResNet50\mbox{--}$\theta_{\mathrm{IN}}$ with $43.7\pm 1.0\%$ macro F1-score) by $11.2$ \gls{pp} (\autoref{fig:results_ranking_rbc}B, \autoref{fig:supp_s3_rbc_lin_probe}). Similarly, at $1\%$ of $\mathcal{D}_{\mathrm{ann}}$, ResNet50 pretrained with \gls{bt} followed by linear probing improved the supervised baseline (ResNet50\mbox{--}$\theta_{\mathrm{IN}}$ with $79.5\pm 0.9\%$ macro F1-score) by $4.3$ \gls{pp}. Furthermore, we notice that representation learning via self-supervised pretraining from random initializations $\theta_{\mathrm{Rnd}}$ yields more improvements than ImageNet initializations $\theta_{\mathrm{IN}}$. Specifically, ConvNeXtV2\mbox{--}T\mbox{--}$\theta_{\mathrm{Rnd}}$ pretrained with MoCo-v3 consistently outperformed the ResNet50\mbox{--}$\theta_{\mathrm{IN}}$ supervised baseline under both XGBoost and linear probing evaluation protocols across almost all data fractions (\autoref{fig:results_ranking_rbc}B, \autoref{fig:supp_s3_rbc_xgboost}, \autoref{fig:supp_s3_rbc_lin_probe}). This superior performance was stable for the XGBoost evaluation protocol, and at the $1\%$, $5\%$, $10\%$, $50\%$, and $100\%$ annotated fractions for the linear probing evaluation protocol. At the $25\%$, MoCo-v2 emerged as the top-performing configuration, while \gls{bt} excelled in the highly resource-constrained $0.1\%$ regime. This variation suggests that while MoCo-v3 provides the most robust representations overall, alternative \gls{ssl} objectives can offer specialized advantages under specific annotation scarcity constraints.

{\setlength{\tabcolsep}{1pt}   
\renewcommand{\arraystretch}{0.7} 
\begin{table*}[!t]
\centering
\caption{Evaluation of the \gls{ssl} methods on the fully annotated \gls{rbc} dataset: Test set macro F1-scores (mean $\pm$ std) for \gls{knn} evaluation, linear probing, and XGBoost. $\theta_{\mathrm{IN}}$ and $\theta_{\mathrm{Rnd}}$ denote initialization from ImageNet-supervised weights and random initialization, respectively. For each evaluation protocol and weight initialization, the best score per backbone is \underline{underlined}, and the best score across all backbones is highlighted in \textbf{bold}.\label{tab:res:ssl:RBC_DS:f1_knn_lp_ft_xgb_split}}
\begin{threeparttable}
\begin{tabular*}{\textwidth}{@{\extracolsep\fill} l l c c c c c c @{\extracolsep\fill}}
\toprule
\textbf{Backbone} & \textbf{Method}
  & \multicolumn{2}{c}{\textbf{\gls{knn}}}
  & \multicolumn{2}{c}{\textbf{Linear probing}}
  & \multicolumn{2}{c}{\textbf{XGBoost}} \\
\cmidrule(lr){3-4} \cmidrule(lr){5-6} \cmidrule(lr){7-8}
&
  & $\mathbf{\theta_{Rnd}}$ & $\mathbf{\theta_{IN}}$
  & $\mathbf{\theta_{Rnd}}$ & $\mathbf{\theta_{IN}}$
  & $\mathbf{\theta_{Rnd}}$ & $\mathbf{\theta_{IN}}$ \\
\midrule
\multirow{7}{*}{ResNet50}
  & SimCLR  & $0.579\pm0.037$ & $0.796\pm0.002$ & $0.752\pm0.018$ & $0.873\pm0.008$ & $0.698\pm0.013$ & $0.891\pm0.008$ \\
  & MoCo-v2  & $0.461\pm0.026$ & $0.829\pm0.020$ & $0.651\pm0.001$ & $0.865\pm0.010$ & $0.641\pm0.026$ & $0.901\pm0.014$ \\
  & MoCo-v3  & $0.481\pm0.032$ & $\underline{\mathbf{0.843\pm0.025}}$ & $0.585\pm0.175$ & $0.857\pm0.013$ & $0.663\pm0.055$ & $0.902\pm0.004$ \\
  & DINO    & $0.558\pm0.006$ & $0.701\pm0.027$ & $0.754\pm0.030$ & $0.829\pm0.003$ & $\underline{0.755\pm0.011}$ & $0.826\pm0.013$ \\
  & \gls{bt}      & $\underline{0.593\pm0.006}$ & $0.814\pm0.009$ & $\underline{0.774\pm0.014}$ & $\underline{0.891\pm0.012}$ & $0.722\pm0.007$ & $\underline{\mathbf{0.904\pm0.004}}$ \\
  & \gls{byol}    & $0.459\pm0.000$ & $0.547\pm0.000$ & $0.605\pm0.000$ & $0.637\pm0.000$ & $0.647\pm0.000$ & $0.665\pm0.000$ \\
\midrule
\multirow{7}{*}{ConvNeXtV1\mbox{--}T}
  & SimCLR  & $0.134\pm0.028$ & $0.789\pm0.015$ & $\underline{0.325\pm0.114}$ & $0.892\pm0.008$ & $0.432\pm0.123$ & $0.879\pm0.009$ \\
  & MoCo-v2  & $\underline{0.261\pm0.046}$ & $0.790\pm0.004$ & $0.263\pm0.060$ & $\underline{0.903\pm0.007}$ & $0.356\pm0.075$ & $\underline{0.896\pm0.001}$ \\
  & MoCo-v3  & $0.192\pm0.007$ & $\underline{0.791\pm0.027}$ & $0.179\pm0.038$ & $0.900\pm0.011$ & $0.250\pm0.017$ & $0.880\pm0.014$ \\
  & DINO    & $0.183\pm0.000$ & $0.678\pm0.000$ & $0.073\pm0.000$ & $0.828\pm0.000$ & $0.228\pm0.000$ & $0.791\pm0.000$ \\
  & \gls{bt}      & $0.246\pm0.085$ & $0.726\pm0.014$ & $0.311\pm0.031$ & $0.857\pm0.021$ & $\underline{0.434\pm0.056}$ & $0.841\pm0.015$ \\
  & \gls{byol}    & $0.185\pm0.000$ & $0.469\pm0.344$ & $0.124\pm0.000$ & $0.423\pm0.527$ & $0.249\pm0.000$ & $0.469\pm0.344$ \\
\midrule
\multirow{7}{*}{ConvNeXtV2\mbox{--}T}
  & SimCLR  & $0.773\pm0.014$ & $0.794\pm0.007$ & $0.891\pm0.012$ & $0.902\pm0.008$ & $0.871\pm0.004$ & $\underline{0.889\pm0.007}$ \\
  & MoCo-v2  & $0.748\pm0.000$ & $0.785\pm0.007$ & $0.903\pm0.007$ & $\underline{\mathbf{0.910\pm0.008}}$ & $0.863\pm0.010$ & $0.875\pm0.001$ \\
  & MoCo-v3  & $\underline{\mathbf{0.828\pm0.007}}$ & $0.753\pm0.014$ & $\underline{\mathbf{0.905\pm0.000}}$ & $0.884\pm0.000$ & $\underline{\mathbf{0.904\pm0.000}}$ & $0.853\pm0.007$ \\
  & DINO    & $0.185\pm0.000$ & $0.590\pm0.143$ & $0.135\pm0.030$ & $0.694\pm0.175$ & $0.238\pm0.002$ & $0.726\pm0.098$ \\
  & \gls{bt}      & $0.780\pm0.006$ & $\underline{0.817\pm0.007}$ & $0.893\pm0.003$ & $0.898\pm0.008$ & $0.862\pm0.008$ & $0.886\pm0.001$ \\
  & \gls{byol}    & $0.098\pm0.007$ & $0.094\pm0.008$ & $0.127\pm0.007$ & $0.137\pm0.030$ & $0.305\pm0.001$ & $0.230\pm0.035$ \\
\midrule
\multirow{7}{*}{\gls{vit}\mbox{--}S/16}
  & SimCLR  & $0.504\pm0.004$ & $0.630\pm0.016$ & $\underline{0.675\pm0.036}$ & $\underline{0.806\pm0.028}$ & $0.612\pm0.022$ & $0.755\pm0.014$ \\
  & MoCo-v2  & $\underline{0.508\pm0.020}$ & $0.612\pm0.019$ & $0.657\pm0.030$ & $0.768\pm0.023$ & $\underline{0.623\pm0.034}$ & $0.722\pm0.003$ \\
  & MoCo-v3  & $0.488\pm0.011$ & $0.610\pm0.017$ & $0.639\pm0.009$ & $0.774\pm0.001$ & $0.604\pm0.012$ & $0.720\pm0.011$ \\
  & DINO    & $0.413\pm0.015$ & $\underline{0.707\pm0.062}$ & $0.492\pm0.042$ & $0.793\pm0.039$ & $0.522\pm0.009$ & $\underline{0.788\pm0.049}$ \\
  & \gls{bt}      & $0.474\pm0.032$ & $0.645\pm0.004$ & $0.632\pm0.002$ & $0.800\pm0.002$ & $0.587\pm0.023$ & $0.748\pm0.018$ \\
  & \gls{byol}    & $0.330\pm0.020$ & $0.576\pm0.126$ & $0.321\pm0.059$ & $0.699\pm0.113$ & $0.411\pm0.006$ & $0.676\pm0.101$ \\
\midrule
\multicolumn{2}{l}{\textbf{Overall median}}
  & $0.468$ & $0.716$ & $0.618$ & $0.843$ & $0.608$ & $0.833$ \\
\bottomrule
\end{tabular*}
\end{threeparttable}
\end{table*}
}

\subsection*{Annotation-Efficient Finetuning: Matching Baselines with Only 5\% annotated Data}\label{subsec:res_finetuning_rbc}

To assess the potential performance gains from finetuning, we selected the top-performing backbone and \gls{ssl} method combinations identified during our initial evaluation for downstream finetuning. These optimal configurations included: ConvNeXtV2\mbox{--}T\mbox{--}$\theta_{\mathrm{Rnd}}$ + MoCo-v3, ResNet50\mbox{--}$\theta_{\mathrm{IN}}$ + MoCo-v3, ConvNeXtV2\mbox{--}T\mbox{--}$\theta_{\mathrm{IN}}$ + MoCo-v2, and ResNet50\mbox{--}$\theta_{\mathrm{Rnd}}$ + \gls{bt} (\autoref{tab:res:ssl:RBC_DS:f1_knn_lp_ft_xgb_split}, \autoref{fig:results_ranking_rbc}D). Every finetuned \gls{ssl} combination consistently outperformed the supervised baseline ConvNeXtV2\mbox{--}T\mbox{--}$\theta_{\mathrm{IN}}$, demonstrating the utility of representation learning. The most substantial gains were concentrated in extreme low-data regimes. At $0.1\%$ and $1\%$ annotated fractions, we observed remarkable absolute test set macro F1-score increases of $45.7$ \gls{pp} and $45.9$ \gls{pp}, respectively, between the supervised baselines and the corresponding top-performing \gls{ssl} combinations (\autoref{fig:results_ranking_rbc}D, E).

A granular, class-specific diagnostic analysis revealed that this low-data superiority was driven by an improvement in resolving highly challenging cellular phenotypes (\autoref{fig:results_ranking_rbc}E). Specifically, at $0.1\%$ $\mathcal{D}_{\mathrm{ann}}$, the detection of crenated discs increased by $62.2$ \gls{pp} in F1-score compared to the baseline. Similarly, at $1\%$ of $\mathcal{D}_{\mathrm{ann}}$, we observed an F1-score enhancement of $57.0$ \gls{pp} for crenated spheroids. For the remaining annotated fractions, the class specific performance gains were mostly driven by a drastic reduction in error rates for the "undecidable" cell category (\autoref{fig:results_ranking_rbc}E). This category poses the steepest challenge due to its high intra-class variance, containing heterogeneous cell structures, edge-cases, and poorly defined morphological signatures \citep{doan_objective_2020}. 

The fine-grained performance trajectories across annotated fractions further exposed how architectural properties interact with self-supervision methods: First, at $0.1\%$ $\mathcal{D}_{\mathrm{ann}}$, the overall best-performing combination was ResNet50\mbox{--}$\theta_{\mathrm{Rnd}}$ + \gls{bt}. Second, the vulnerability of ConvNeXtV2\mbox{--}T\mbox{--}$\theta_{\mathrm{IN}}$ at very low data regimes observed in the supervised baselines (\autoref{fig:supervised_baselines}, \autoref{fig:supp_supp_baselines}B) was mitigated but did not completely disappear. From $5\%$ annotated fraction onward, the pretrained \gls{ssl} configurations stabilized, and ConvNeXtV2\mbox{--}T\mbox{--}$\theta_{\mathrm{IN}}$ + MoCo-v2 emerged as best performer with a macro F1-score of $92.3\% \pm1.0$, equaling the best supervised baseline trained on full $\mathcal{D}_{\mathrm{ann}}$ (ConvNeXtV2\mbox{--}T\mbox{--}$\theta_{\mathrm{IN}}$ with $92.2 \pm 1.1\%$ macro F1-score) (\autoref{tab:res:supervised:baelines}). This sudden inflection point underscores the synergistic potential of pairing advanced backbone feature extractors with optimized \gls{ssl} methods, ultimately enabling highly efficient downstream annotation utilization.

\subsection*{Qualitatively Mapping the Self-Supervised Latent Space}\label{subsec:umpas}
\begin{figure*}[!htb]
\centerline{\includegraphics[width=\linewidth]{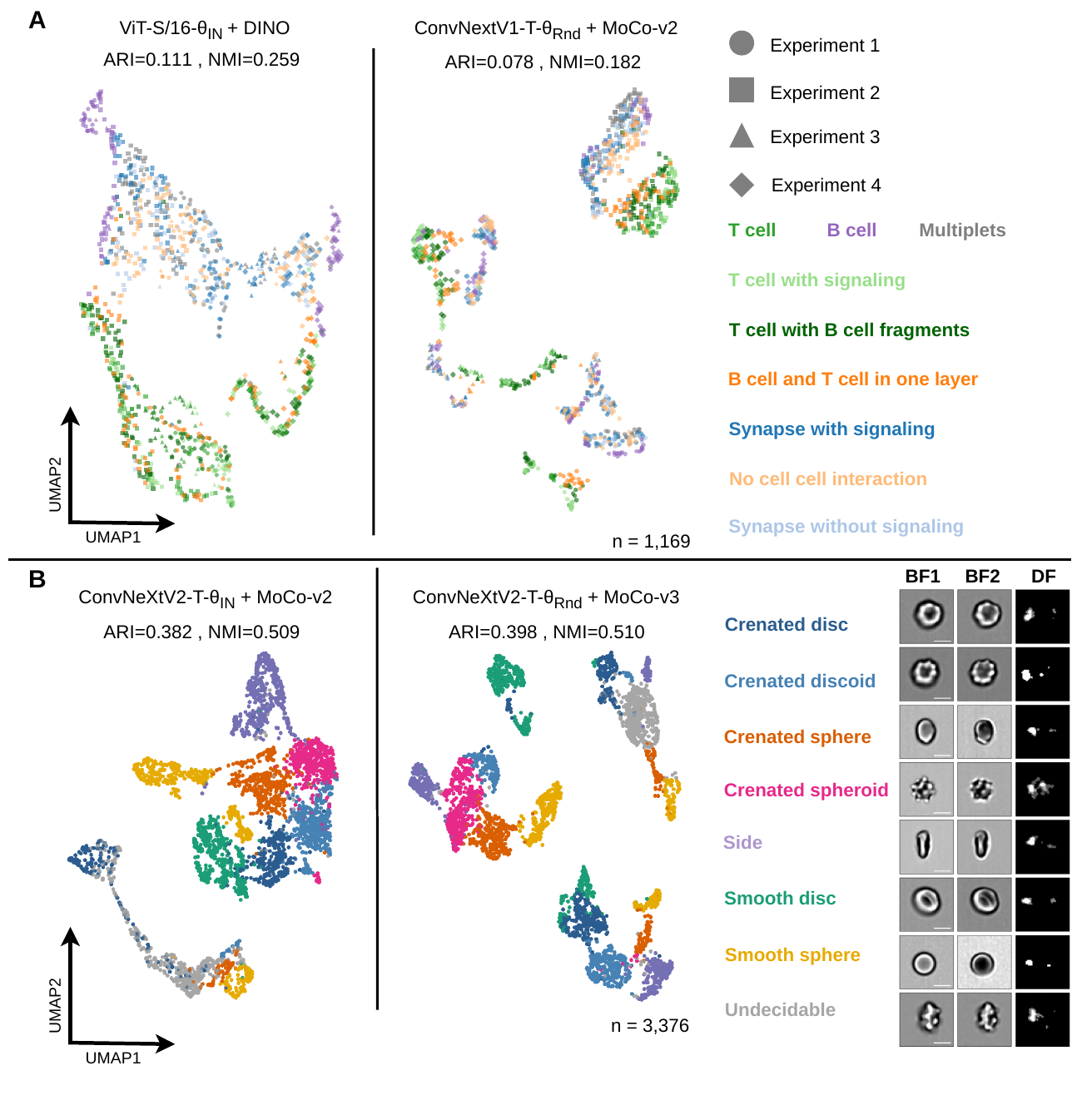}}
\caption{Latent space visualization of the pretrained backbones on full $\mathcal{D}_{\mathrm{una}}$ for the \textbf{(A)} \gls{isf}\mbox{--}5CH and \textbf{(B)} \gls{rbc} datasets. We selected for each initialization strategy ($\theta_{\mathrm{IN}}$ or $\theta_{\mathrm{Rnd}}$) the best \gls{ssl} combination after linear probing evaluation and color-coded test samples with annotations. \gls{ari} stands for \acrlong{ari} and \gls{nmi} for \acrlong{nmi}. Scale bars in illustratives images of \textbf{(B)} represent $5\mu m$. Experiments in \textbf{(A)} represent the individual experiments as described in~\citep{shetab_boushehri_explainable_2023, essig_imaging_2022}.\label{fig:results_umaps}}
\end{figure*}

To qualitatively assess the structure and semantics of the learned representations, we visualized the high-dimensional latent space of our best-performing \gls{ssl} combinations for both $\theta_{\mathrm{Rnd}}$ and $\theta_{\mathrm{IN}}$ under linear probing using a 2D-\gls{umap} (\autoref{fig:results_umaps}). Because low-dimensional projections distort distances and cannot support quantitative claims, we clustered the embeddings with $k$-means and scored the resulting partition against the annotations using the \gls{ari} and the \gls{nmi} (\autoref{tab:supp:cluster_structure}). On the \gls{isf}\mbox{--}5CH dataset, we additionally clustered the embeddings of each experiment separately and compared the resulting within experiment scores against a control in which images were assigned at random to groups of identical size (\autoref{tab:supp:cluster_structure}). 

{\setlength{\tabcolsep}{4pt}   
\renewcommand{\arraystretch}{0.95} 
\begin{table}[!htb]
\centering
\caption{Assessing the cluster structure of the learned representations on the \gls{isf}\mbox{--}5CH dataset with \gls{ari} and \gls{nmi}. Unsupervised $k$-means clustering of embeddings with $k$ set to the number of groups and $n$ representing the number of images. Class within experiment is computed separately for each experiment; the weighted mean averages over experiments proportionally to the number of images. The random control repeats the within-experiment analysis on size-matched random groupings ($10$ repetitions). \label{tab:supp:cluster_structure}}
\begin{center}
\begin{tabular*}{\textwidth}{@{\extracolsep\fill}lccccccc@{\extracolsep\fill}}
\toprule
 & & & \multicolumn{2}{c}{\textbf{\mbox{ConvNeXtV1\mbox{--}T\mbox{--}$\theta_{\mathrm{Rnd}}$ + MoCo-v2}}}&\multicolumn{2}{c}{\textbf{\mbox{\gls{vit}\mbox{--}S/16\mbox{--}$\theta_{\mathrm{IN}}$ + DINO}}}\\
\cmidrule{4-5}\cmidrule{6-7}
\textbf{Grouping} & \textbf{$n$} & \textbf{$k$} & \textbf{ARI} & \textbf{NMI} & \textbf{ARI} & \textbf{NMI} \\
\midrule
Annotated class & $1,169$ & $9$ & $0.078$ & $0.182$ & $0.111$ & $0.259$\\
Experiment & $1,169$ & $4$ & $0.811$ & $0.792$ & $0.212$ & $0.349$\\
\midrule
\multicolumn{7}{@{}l}{\textit{Class within experiment}} \\
\quad Experiment 1 & $252$ & $9$ & $0.188$ & $0.353$ & $0.201$ & $0.413$\\
\quad Experiment 2 & $139$ & $9$ & $0.160$ & $0.398$ & $0.306$ & $0.533$\\
\quad Experiment 3 & $486$ & $9$ & $0.146$ & $0.294$ & $0.170$ & $0.347$\\
\quad Experiment 4 & $292$ & $9$ & $0.154$ & $0.353$ & $0.219$ & $0.425$\\
\quad Weighted mean & $1,169$ & $9$ & $0.159$ & $0.334$ & $0.205$ & $0.403$\\
\quad Random control & $1,169$ & $9$ & $0.086$ & $0.233$ & $0.111$ & $0.288$\\
\bottomrule
\end{tabular*}
\end{center}
\end{table}
}

A detailed examination of the individual datasets reveals specific topological behaviors that further illustrate the clinical utility and robustness of self-supervised representations:
\begin{itemize}
    \item \textbf{Experimental Batch Effects vs. Class Clustering}: On the \gls{isf}\mbox{--}5CH dataset, the latent space projection for both \gls{ssl} combinations exhibit prominent batch effects governed by the distinct experiments as described in~\citep{shetab_boushehri_explainable_2023, essig_imaging_2022} (\autoref{fig:results_umaps}A). This is confirmed when looking at the clustering results on annotations and experiments (\autoref{tab:supp:cluster_structure}). When grouping by annotated classes, the best \gls{ssl} combination, ConvNeXtV1\mbox{--}T\mbox{--}$\theta_{\mathrm{Rnd}}$ + MoCo-v2  achieves an \gls{ari} of $0.078$ and an \gls{nmi} of $0.182$. These values rise to $0.811$ and $0.792$ respectively when grouping by experiments. This clustering behavior, typical of multi-batch biological assays with varying experimental conditions, could superficially appear to degrade semantic organization. However, when looking closer into individual batch sub-clusters, we observe a coherent internal organization consisting of boundaries between immunological classes. For example, we notice a separation between T cell images and others (\autoref{fig:results_umaps}A). This is further noticeable in ConvNeXtV1\mbox{--}T\mbox{--}$\theta_{\mathrm{Rnd}}$ + MoCo-v2, with higher \gls{ari} and \gls{nmi} within individual experiments (\autoref{tab:supp:cluster_structure}), which, averaged (\gls{ari} $= 0.159$, \gls{nmi} $= 0.334$), surpassed the original grouping by annotated classes and the random control.  More importantly, we notice the aforementioned complexity of the class "B cell and T cell in one layer" (B cell and T cell completely overlapping as shown in \autoref{fig:experiments_datasets}C), which correlates with the substantial improvements noticed in finetuning (\autoref{fig:results_ranking_isf}E). Additionally, the \gls{vit}\mbox{--}S/16\mbox{--}$\theta_{\mathrm{IN}}$ + DINO displays a superior capacity to mitigate these persistent batch variations (\autoref{tab:supp:cluster_structure}). This batch-reconciliation capability is particularly visible in the better clustering based on annotated classes compared to ConvNeXtV1\mbox{--}T\mbox{--}$\theta_{\mathrm{Rnd}}$ + MoCo-v2 (\autoref{fig:results_umaps}A). This may suggest that the \gls{vit}\mbox{--}S/16\mbox{--}$\theta_{\mathrm{IN}}$ + DINO successfully leverages subtle visual features beyond the fluorescent signal to achieve precise B-cell isolation, demonstrating that out-of-domain pretraining enables the network to look past surface-marker ambiguities to capture deep, structural cellular lineages.
    \item \textbf{Pretext-Guided Annotation and Morphology Mapping}: On the \gls{rbc} dataset, the latent spaces of both models showed several clusters. While the randomly initialized latent space ConvNeXtV2\mbox{--}T\mbox{--}$\theta_{\mathrm{Rnd}}$ + MoCo-v3 struggled with diffuse and noisy clusters, the ImageNet-pretrained configuration ConvNeXtV2\mbox{--}T\mbox{--}$\theta_{\mathrm{IN}}$ + MoCo-v2 established better segregated topology. Although this pretrained latent space displays some local neighborhood overlap, we observe that it successfully disentangles visually similar morphological variants (\autoref{fig:results_umaps}B), specifically resolving crenated spheres from crenated spheroids. This structural division directly aligns with the class-specific performance gains we obtained during finetuning (\autoref{fig:results_ranking_rbc}E). Furthermore, the two distinct clusters of smooth discs, crenated discoids, and side cells present in ConvNeXtV2\mbox{--}T\mbox{--}$\theta_{\mathrm{Rnd}}$ + MoCo-v3  are merged under ConvNeXtV2\mbox{--}T\mbox{--}$\theta_{\mathrm{IN}}$ + MoCo-v2. Of great interest is the distribution of the "undecidable" class. Rather than behaving as an unstructured, random noise overlay, the "undecidable" cells organize into tight, localized sub-pockets embedded directly within the boundaries of major, well-defined phenotypic clusters. This localized segregation suggests that these ambiguous cells share strong, continuous morphological features with their neighboring defined classes. Consequently, these structural mappings serve as a valuable visual taxonomy tool, helping pathologists and experimentalists pinpoint exactly which biological phenotypes these uncertain structures are most morphologically related to, thereby accelerating downstream annotator consensus and streamlining semi-supervised annotating pipelines.

\end{itemize}

Taken together, these latent space topologies confirm that qualitative feature clustering directly correlates with our quantitative downstream evaluations. They demonstrate that while target-domain self-supervised pretraining ($\theta_{\mathrm{Rnd}}$) refines specialized features, the structural priors inherited from natural image pretraining ($\theta_{\mathrm{IN}}$) could improve the fine-grained morphological boundaries that ultimately drive high-precision downstream analysis.

\section*{Discussion and Guidelines}\label{sec:discussion}

Annotation remains the principal bottleneck in \gls{ifc} analysis, where acquisition routinely outpaces expert annotating by orders of magnitude. We benchmarked six \gls{ssl} objectives across four backbones and two initialization strategies on two \gls{ifc} datasets, holding preprocessing and pretraining scale fixed so that comparisons isolate the objective and the initialization. Self-supervised pretraining reduced annotation requirements substantially and successfully disentangles complex biological phenotypes. 

\subsection*{Weights Initialization should be tailored to Task Complexity}\label{subsec:discussion_init}

Despite some studies suggesting that backbone initialization with ImageNet-supervised weights consistently enhances biomedical image representations~\citep{wang2023real}, recent evaluations have begun to reveal its limitations, specifically showing that supervised natural image priors do not consistently extend to complex downstream tasks and can even degrade robustness under certain \gls{ssl} combinations~\citep{bundele2024evaluating}. Our results solidify these findings. Crucially, our extensive evaluation reveals that the optimal backbone initialization strategy is highly domain-dependent rather than universal: While fine-grained geometric features in the \gls{rbc} dataset benefit significantly from natural image weights ($\theta_{\mathrm{IN}}$) (\autoref{tab:res:ssl:RBC_DS:f1_knn_lp_ft_xgb_split}, \autoref{fig:supp_s3_rbc_knn}, \autoref{fig:supp_s3_rbc_lin_probe}, \autoref{fig:supp_s3_rbc_xgboost}), more complex cell structural phenotypes from the \gls{isf}\mbox{--}5CH dataset are better resolved via in-domain self-supervision from random initialization ($\theta_{\mathrm{Rnd}}$) (\autoref{tab:res:ssl:ISF-5CH_DS:f1_knn_lp_xgb}, \autoref{fig:supp_s3_isf_knn}, \autoref{fig:supp_s3_isf_lin_probe}, \autoref{fig:supp_s3_isf_xgboost}). This pattern was present for each annotation scarcity level, helping address \textbf{RQ1, RQ2} and providing a proposal for optimizing representation quality under raw imaging constraints, which we leverage here to establish a concrete guideline for backbone initialization.

\subsection*{A Strong Recipe for Learning Linearly Separable Features: MoCo-v2 + ConvNeXt}\label{subsec:discussion_combi}

While claiming, in response to \textbf{RQ1}, that a singular self-supervised paradigm can dominate diverse biological imaging tasks may appear ambitious, our empirical evidence systematically validates this reality. Across our benchmarking experiments under the strict linear probing protocol, the combination of MoCo-v2 with a modern ConvNeXt backbone emerged as the most robust baseline (\autoref{tab:res:ssl:ISF-5CH_DS:f1_knn_lp_xgb}, \autoref{tab:res:ssl:RBC_DS:f1_knn_lp_ft_xgb_split}), thus aligning with recent applications of \gls{ssl} on biomedical imaging~\citep{bundele2024evaluating, Huang2026-if}. On the \gls{rbc} dataset, this peak configuration was achieved using a ConvNeXtV2\mbox{--}T backbone with $\theta_{\mathrm{IN}}$. Conversely, on the \gls{isf}\mbox{--}5CH dataset, the optimal pipeline shifted to a ConvNeXtV1\mbox{--}T backbone trained completely from scratch. These results additionally yield a clear backbone architectural prescription: Practitioners should prioritize modern ConvNeXt or \gls{vit}\mbox{--}S/16 backbones over classical residual networks. In our evaluations, the standard ResNet50 showed fluctuations in performance. On the one hand, ResNet50 combined with contrastive methods MoCo-v2 or MoCo-v3 was effective. On the other hand, its performance was worse on the asymmetric methods DINO and \gls{byol}, particularly on the \gls{isf}\mbox{--}5CH dataset. This shows ResNet50's dependency on \gls{ssl} objectives, suggesting that large-kernel spatial mixing of ConvNeXt or the global receptive fields of \glspl{vit} are likely to be less dependent on \gls{ssl} objectives and better at capturing the fine-grained hierarchical textures of cellular imaging. Similar to other benchmarks~\citep{baharoon2023evaluating}, we found that some specific \gls{ssl} combinations are highly compatible. For example, pairing a \gls{vit}\mbox{--}S/16 with the DINO framework yielded a potent synergy that achieves competitive results across both evaluation datasets. 

Our hierarchical benchmarking and ranking showed a transition under extreme downstream annotation scarcity, where specialized self-supervised paradigms revealed their algorithmic advantages. On the large \gls{isf}\mbox{--}5CH dataset, DINO exhibits on modern backbones like ConvNeXts and \gls{vit}\mbox{--}S/16 impressive performances at ultra-low data annotation fractions (\autoref{fig:results_ranking_isf}B), showcasing the strength of its local to global self-distillation framework in clustering complex cellular features using minimal supervision~\citep{DBLP:conf/iccv/CaronTMJMBJ21}. Conversely, on the relatively medium scale \gls{rbc} dataset, \gls{bt} dominates the ultra-low regime. In contrast, \gls{byol} exhibits severe performance fluctuations and consistently underperforms across both datasets. This deficit aligns with established literature highlighting its sensitivity to the hyperparameters~\citep{pmlr-v139-tian21a} and suggests that it is susceptible to partial dimensional collapse when exposed to the highly homogeneous and repetitive background profiles of \gls{ifc}.

\subsection*{Finetuning enables Annotation Efficiency}\label{subsec:discussion_finetuning}
In our experiments to address \textbf{RQ2}, self-supervised pretraining on unannotated data translated into large annotation efficiency during downstream finetuning. For example, with roughly $1,000$ annotated samples on the \gls{isf}\mbox{--}5CH dataset, a simple finetune performance exceeded the fully supervised baselines (\autoref{fig:results_ranking_isf}D). On the \gls{rbc} dataset, we matched the $100\%$ annotation fraction baseline using only $5\%$ of the annotations (\autoref{fig:results_ranking_rbc}D). These results imply that, when abundant unannotated \gls{ifc} data are available, practitioners should pretrain encoders with \gls{ssl} and then finetune on a very small number of annotated examples: This workflow reliably enables highly annotation-efficient learning, as it held on both datasets despite their differences in cell type, marker panel, class definition and size, which suggests it is not specific to a single assay. We did not, however, evaluate transfer of pretrained encoders between independent \gls{ifc} studies. Because acquisition panels in \gls{ifc} are experiment-specific, differing in the number, identity and intensity scaling of channels, encoders pretrained on one panel cannot be applied directly to another without channel adaptation. The backbones released here are therefore intended as initialization for studies using comparable channel configurations, and cross-study transfer, including channel-adaptive pretraining, remains an open question.

\subsection*{Practical Implementation and Normalization Pitfalls}\label{subsec:discussion_details}
Our experiments highlighted the utility of qualitative performance assessment (\autoref{fig:results_umaps}).   

For example, we observed on the \gls{isf} dataset that the learned representations were clustered by experiment, suggesting the presence of experiment-level batch effects. However, boundaries and class-specific clusters were still visible in the individual experiment clusters. This serves as strong recommendation for characterizing the latent space qualitatively, as it may provide more insights on the features learned during pretraining.

Our findings suggest that engineering complex domain-specific preprocessing pipelines may yield diminishing returns. Specifically, we advise against introducing cytometric channel scales like the Logicle transformation~\citep{parks_new_2006, moore_update_2012}. While standard in classical flow cytometry analysis, integrating Logicle scaling into visual deep learning pipelines fails to bring any downstream performance improvements over simple z-score normalization. 

Additionally, while we did not evaluate channel importance and/or reduction strategies in this study, seminal deep learning work in \gls{ifc} has systematically demonstrated that omitting channels or restricting models to single modalities discards highly complementary physical and spatial features, leading to a marked degradation in performance~\citep{boushehri2024pxpermute}. We therefore advise practitioners to avoid restricting themselves to a single modality unless justified.

\subsection*{Limitations and Future Work}\label{subsec:discussion_future_work}

While this study establishes baselines for representation learning in \gls{ifc}, several limitations should be noted.

Regarding scope, our benchmarking framework covers discriminative \gls{ssl} only; generative and reconstruction-based approaches, such as masked autoencoders and diffusion-based features, were not evaluated. We likewise did not investigate explicit batch-correction techniques to mitigate the multi-run domain shifts reported above, did not ablate domain-specific transformations, and did not quantify the contribution of individual imaging channels. Each of these would require a further multiplication of an already large configuration grid. We also only used z-score normalization during our benchmarking and therefore cannot conclude if this is optimal. Ablating several techniques could be a promising direction. 

Regarding the evidence, our conclusions rest on two \gls{ifc} datasets. Although these differ in cell type, marker panel, class definition and size, two datasets are a narrow basis for claims about \gls{ifc} in general. Furthermore, each configuration was trained with three seeds, which is sufficient to report the observed variability but limits the resolution at which closely ranked
methods can be separated. Addressing these points, in particular through evaluation on additional independent \gls{ifc} cohorts, remains a promising direction for future work.

\section*{Methods}\label{sec:methods}

Our modular pipeline illustrated in \autoref{fig:method} provides a controlled environment to isolate the factors contributing to representation quality in multispectral \gls{ifc} data. The following sections detail our methodology in a bottom-up approach: Beginning with the selection and preprocessing of publicly available datasets, followed by the architectural configurations of the backbones, the implementation of diverse \gls{ssl} paradigms and finally the standardized evaluation protocols used to measure downstream performance.

\subsection*{Data Preparation}\label{subsec:meth_data}

At this stage, the raw multispectral \gls{ifc} data are ingested as a collection of multichannel tensors, where each sample $x \in \mathbb{R}^{C \times H \times W}$ with $C$ the number of channels, $H$ the image height and $W$ the image width. To facilitate benchmarking, the dataset $\mathcal{D}$ is partitioned into two disjoint subsets: an unannotated set $\mathcal{D}_{\mathrm{una}}$ for self-supervised pretraining and a annotated set $\mathcal{D}_{\mathrm{ann}}$ reserved for downstream supervised evaluation. 

Our framework incorporates a channel selection module to isolate specific biomarkers and a preprocessing suite that supports domain-specific transformations (e.g., Logicle \citep{moore_update_2012, parks_new_2006} or Hyperlog \citep{bagwell_hyperlog-flexible_2005} scaling). While the framework allows for these specialized preprocessing techniques, they were excluded from the primary benchmarking experiments, specifically for those consisting solely of brightfield $\mathrm{BF}$ and/or darkfield $\mathrm{DF}$ imagery. We reserved the exploration of these domain-specific techniques for our ablation studies, focusing instead on standard channel-wise z-score normalization on the full dataset and spatial resizing ($224 \times 224$) to ensure a uniform input format across all \gls{ssl} paradigms. We selected for our benchmarking two publicly available \gls{ifc} datasets:

\begin{itemize}
\item \gls{rbc} dataset~\citep{doan_objective_2020}, a publicly available medium-scale \gls{ifc} dataset of single-cell red blood cell images. The dataset consists of an unannotated subset of $245,085$ images from three bags on three different days each~\citep{doan_objective_2020, Doan2020_unlabeled}, and a annotated subset of $70,331$ samples classified in seven expert-defined morphological categories (Crenated disc, Crenated discoid, Crenated sphere, Crenated spheroid, Side, Smooth disc, Smooth sphere) and one undecidable category~\citep{doan_objective_2020}. Each sample contains two brightfield channels ($\mathrm{BF1}$ and $\mathrm{BF2}$) and one darkfield channel ($\mathrm{DF}$). For this benchmark, we utilize all available channels $\mathrm{BF1}$, $\mathrm{BF2}$, and $\mathrm{DF}$ to ensure a robust evaluation of representation capacity. According to \citep{doan_objective_2020}, samples were acquired using the Amnis ImageStreamX Mark II and preprocessed with the corresponding software IDEAS. Dataset statistics and representative class examples are shown in \autoref{fig:experiments_datasets}A and B. The splits were taken from the published data in the following manner: The folder named "Training Test 2 Canada Swiss" was used for the training, "Test 3 4 Holdout Canada Swiss" was used for the validation, and "Test 1 Consensus ground truth" was used for test. For this dataset, the unannotated partition set $\mathcal{D}_{\mathrm{una}}$ consisted of four-channel (an additional $\mathrm{DF}$ to the three channels aforementioned in the annotated subset) images of varying spatial dimensions, whereas the annotated subset $\mathcal{D}_{\mathrm{ann}}$ comprised uniform $48 \times 48$ pixel images restricted to three aforementioned channels. To maintain channel consistency between the pretraining and evaluation phases, pretraining on $\mathcal{D}_{\mathrm{una}}$ was restricted to these same three channels.
\item \gls{isf} dataset, one of the largest publicly available \gls{ifc} dataset~\citep{shetab_boushehri_explainable_2023, essig_imaging_2022}. It profiles immunological synapse formation between human memory CD4$^+$ T cells and \gls{ebv} B lymphoblastoid cells (B-LCLs)~\citep{shetab_boushehri_explainable_2023, essig_imaging_2022}. T cells from nine donors were co-cultured with \gls{sea}-pulsed or untreated B-LCLs and imaged on an Amnis ImageStreamX Mark II across four independent experiments, yielding more than $2.89$ million images~\citep{shetab_boushehri_explainable_2023, essig_imaging_2022} as shown in \autoref{fig:experiments_datasets}D. Each sample contains brightfield ($\mathrm{BF}$) and fluorescence channels ($\mathrm{CD45}$, $\mathrm{Antibody}$, $\mathrm{F\mbox{--}actin}$, $\mathrm{MHCII}$, $\mathrm{CD3}$, $\mathrm{P\mbox{--}CD3\zeta}$) plus a live/dead stain ($\mathrm{L/D}$). A subset of $9,186$ images was manually annotated by an expert immunologist into $17$ classes~\citep{shetab_boushehri_explainable_2023,boushehri2024pxpermute}. For this study, we focus on the nine biologically relevant classes (B cell, T cell, T cell with signaling, B cell and T cell in one layer, T cell with B cell fragments, Synapse without signaling, Synapse with signaling, No cell cell interaction, and Multiplets) as illustrated in \autoref{fig:experiments_datasets}C, yielding a filtered set of $5,848$ samples. The ignored classes consisted mostly of undesired samples (Unknown, Dead cell, Out of focus, Unfocused synapses, Cropped images, No synapse, Unfocused cells, and CD3+MHCII+cells).  The obtained subset is split for the downstream tasks into training, validation, and test with a ratio of $64$-$16$-$20$.
\end{itemize}
All images across both the \gls{rbc} and \gls{isf} datasets were acquired at $60\times$ magnification with a native spatial resolution of $0.33$ micrometers per pixel. We follow the original studies and use the same channels for the benchmarking. All images of both datasets were resized to $224 \times 224$.

To benchmark downstream models under varying degrees of annotation scarcity, $\mathcal{D}_{\mathrm{ann}}$ was partitioned into nested subsets representing $1\%$, $5\%$ $10\%$, $25\%$, $50\%$, and $100\%$ of the total available training set. To prevent class imbalance from biasing evaluation at ultra-low regimes (such as $1\%$), all subsets were generated using stratified random sampling to preserve the parental class distributions. We maintained consistent random seeds across experiments to ensure that smaller target subsets (e.g., $1\%$) were strictly nested subsets of larger ones (e.g., $5\%$), isolating data volume as the sole experimental variable.

\subsection*{Self-Supervised Pretraining}\label{subsec:meth_ssl}
As illustrated in \autoref{fig:method}, the second stage of our pipeline is a fully decoupled pretraining module designed to evaluate the synergy between model architecture and self-supervised objectives systematically. The construction of the pretraining environment follows a structured and modular hierarchy. 

\subsubsection*{Backbones}\label{subsubsec:meth_backbones}
To ensure the generalizability of our benchmarking results, we integrate four distinct backbone architectures that represent the core paradigms of modern computer vision. We employ ResNet50~\citep{DBLP:conf/cvpr/HeZRS16} as a convolutional baseline to assess feature extraction via standard residual learning. To contrast this traditional approach with global dependency modeling, we incorporate a \gls{vit}\mbox{--}S/16~\citep{DBLP:conf/iclr/DosovitskiyB0WZ21}, which utilizes self-attention mechanisms to process imagery as sequential patches. Furthermore, we include ConvNeXtV1\mbox{--}T~\citep{DBLP:conf/cvpr/0003MWFDX22} and its successor, ConvNeXtV2\mbox{--}T~\citep{DBLP:conf/cvpr/WooDHC0KX23}, which represent modernized convolutional designs that adopt Transformer-inspired hierarchies, such as larger kernels, while maintaining convolutional efficiency. Critically, we specifically utilize the $Tiny$ and $Small$ variants of ConvNeXts and \gls{vit}\mbox{--}S/16 to maintain a comparable parameter budget across the different architectural families, ensuring that performance variations are attributable to structural differences rather than sheer model capacity. As detailed in \autoref{tab:exp:backbones}, these models span a wide spectrum of depths, and receptive field sizes. Each architecture is evaluated under two initialization regimes: random ($\theta_{\mathrm{Rnd}}$), to measure the raw representation learning on \gls{ifc} data, and ImageNet-pretrained ($\theta_{\mathrm{IN}}$), to quantify the degree of knowledge transfer from natural image domains to the biological landscape.

{\setlength{\tabcolsep}{4pt}   
\renewcommand{\arraystretch}{0.95} 
\begin{table}[!htb]
\centering
\caption{Overview of the benchmarked backbone architectures. Parameter counts expressed in millions.
\label{tab:exp:backbones}}
\begin{center}
\begin{tabular*}{310pt}{@{\extracolsep\fill}lccc@{\extracolsep\fill}}
\toprule
\textbf{Backbone} & \textbf{\# Parameters} & \textbf{Blocks} & \textbf{Kernel/Patch size} \\
\midrule
ResNet50 & $\sim 23.5$ M & $16$ & $3 \times 3$  \\
ConvNeXtV1\mbox{--}T &  $\sim 27.9$ M & $18 $ &  $7 \times 7$ \\
ConvNeXtV2\mbox{--}T &  $\sim 27.9$ M & $18$ &  $7 \times 7$ \\
\gls{vit}\mbox{--}S/16 &  $\sim 21.7$ M & $12$ & $16 \times 16$\\
\bottomrule
\end{tabular*}
\end{center}
\end{table}
}

To address the mismatch between the three-channel input of ImageNet-pretrained models ($C_{in}=3$) and the multispectral nature of \gls{ifc} data ($C_{in}=N$), we utilize a standardized weight-averaging adaptation. Let the pretrained first-layer kernels be $W_{pre} \in \mathbb{R}^{D \times 3 \times K \times K}$ where $D$ is the number of filters, $3$ the RGB depth, and $K$ the kernel size. To initialize the new weight tensor $W_{target} \in \mathbb{R}^{D \times C_{in} \times K \times K}$ without losing the pre-trained feature detectors, we apply channel-wise mean replication. We first compute the average filter across the three RGB channels to create a gray-equivalent filter template $W_{\mu} \in \mathbb{R}^{D \times 1 \times K \times K}$ as indicated in \autoref{eq:grayscale} where $W_{pre}^{(j)}$ is the weight slice for the j-th color channel.

\begin{equation}
W_{\mu} =  \frac{1}{3}\sum_{j=1}^{3}W_{pre}^{(j)}
\label{eq:grayscale}
\end{equation}

We then replicate this mean filter across all $C_{in}$ channels of the target architecture:
$W_{target}^{(i)}=W_{\mu}, i \in \{1, \dots , C_{in}\}$. 

Because this study aims to compare self-supervised objectives and backbones rather than to optimise channel handling, we opted for this simple trick and did not use any of the current specialized methods~\citep{rs18040655, DBLP:conf/nips/CongKMLRHBLE22, 9511403}.

\subsubsection*{Supervised Baseline}\label{subsubsec:meth_sup_baseline}
To contextualize the quality of the learned self-supervised representations, we established a supervised baseline for each dataset, backbone, and initialization strategy. These baseline models are trained under different scarcity levels with weighted cross-entropy loss and weighted sampling, representing, for each scarcity level, the empirical performance baseline. By comparing results obtained from \gls{ssl} against this ceiling, we can quantify the  supervised-\gls{ssl} gap and evaluate how effectively our framework captures the feature space required for biological classification under various annotation scarcity level. The corresponding hyperparameter values are detailed in \autoref{tab:supp:supp_par}.

\subsubsection*{Self-Supervised Learning Methods}\label{subsubsec:meth_ssl_ssl}
Our framework supports a diverse and extensible registry of \gls{ssl} methods, each characterized by its architectural head and loss function. We however focus here on invariance-based \gls{ssl}. This registry therefore incorporates symmetric contrastive approaches like SimCLR~\citep{DBLP:conf/icml/ChenK0H20}, which uses a shared \gls{mlp} projection head to maximize agreement between two augmented views, and momentum-based variants such as MoCo-v2~\citep{DBLP:journals/corr/abs-2003-04297} and MoCo-v3~\citep{DBLP:conf/iccv/ChenXH21}. While MoCo-v2 utilizes a queue of negative samples to decouple contrastive memory from batch size, MoCo-v3 adopts an asymmetric formulation with an additional prediction head and synchronized batch statistics. The framework also encompasses asymmetric teacher-student architectures like \gls{byol}~\citep{DBLP:conf/nips/GrillSATRBDPGAP20}, which relies on a predictor and stop-gradient operations to prevent collapse without negative samples, and DINO~\citep{DBLP:conf/iccv/CaronTMJMBJ21}, which utilizes a multi-crop strategy for robust self-distillation. Furthermore, the redundancy-reduction approach of \gls{bt}~\citep{DBLP:conf/icml/ZbontarJMLD21} is supported to explicitly penalize correlation across embedding dimensions. By mostly preserving these method specific components, our framework enables a fair and systematic evaluation of how diverse self-supervised methods impact feature quality and downstream performance. A detailed explanation of the benchmarked \gls{ssl} methods and their corresponding training hyperparameters is provided in the Supplementary Information (Supplementary Section '\gls{ssl} Methods' and Supplementary \autoref{tab:supp:ssl_hyperparams}).

\subsubsection*{Pretraining Data Augmentation}\label{subsubsec:meth_ssl_aug}
To ensure a rigorous and controlled benchmarking environment, we implement a standardized data augmentation protocol across all evaluated \gls{ssl} methods. For methods following a standard two-view strategy (i.e., SimCLR, MoCo-v2, MoCo-v3, \gls{bt}, and \gls{byol}), images are processed at a global resolution of $224 \times 224$ pixels. For the multi-crop method  DINO, we maintain consistent view configurations, utilizing two global views at $224 \times 224$ along with 6 local crops at $96 \times 96$. The geometric transformation pipeline consists of horizontal and vertical flips, rotations, and translations. Photometric augmentations include Gaussian blur and noise to account for \gls{ifc} optical characteristics. All parameters for these augmentations can be found in \autoref{tab:supp:ssl_aug_params}.

In contrast to standard self-supervised pipelines for natural images, we exclude color jittering and solarization as they can destroy biologically relevant intensity information~\citep{huang2023self}. By maintaining these augmentation parameters constant across all experiments, including synchronized crop scales and noise sigmas, we ensure that performance variations are attributable strictly to the self-supervised paradigms and backbone configurations under study, rather than disparate data-view policies.

\subsection*{Downstream Analysis}\label{subsec:meth_dse}
To evaluate the quality of the learned representations, we follow the established standard in the field~\citep{marks2025closer} and adopt a multi-tiered evaluation protocol across three standard regimes:
\begin{itemize}
    \item \gls{knn} classification on the l2-normalized extracted feature representations from the frozen backbone with $k=5$ and cosine distance as the proximity metric and neighbor contributions weighted uniformly. This non-parametric approach enables evaluating local feature space density.
    \item Linear probing with a multinomial logistic regression classifier trained directly on the z-score-normalized extracted feature representations from the frozen backbone. Optimization was carried out using the Limited-memory Broyden-Fletcher-Goldfarb-Shanno (L-BFGS) solver, configured with $1,000$ maximum iterations to guarantee full gradient convergence across all evaluation partitions.
    \item End-to-end finetuning to determine the maximum achievable performance for each task. All hyperparameters can be found in \autoref{tab:supp:supp_par}.
\end{itemize}
Additionally, we incorporated an XGBoost classifier as a computationally efficient proxy for characterizing non-linear information within the learned representations. Tree-based ensembles like XGBoost are architectural staples in bioimaging and particularly in \gls{ifc}~\citep{shetab_boushehri_explainable_2023}, due to their robustness on tabular feature sets and lower computational overhead compared to end-to-end neural network optimization. XGBoost was applied with the multi-class softmax objective, $500$ boosting rounds and library defaults. Full end-to-end finetuning was performed only on the top-performing \gls{ssl} configurations identified from the evaluation protocols \gls{knn}, linear probing, and XGBoost for each initialization strategy (random vs. ImageNet-pretrained).

Quantitative performance is measured using the macro F1-score, \gls{ari}, and \gls{nmi}, while the qualitative structure of the latent space is visualized through \gls{umap}~\citep{healy2024uniform} embeddings. For \gls{umap}, embeddings were l2-normalized and high-dimensional distances were calculated using cosine similarity, $15$ nearest neighbors and a minimum distance parameter of $0.4$.

\subsection*{Training and Implementation Details}\label{subsec:meth_implementation_details}
All models were developed in PyTorch 2.7 and trained on NVIDIA A100 GPUs with 40 GB VRAM. We maintained a fixed seeding strategy to ensure reproducibility of the stochastic components. Final results represent the aggregate of three runs, reported via mean and standard deviation. Further implementation details can be found in the supplementary materials. 

To adapt the self-supervised pretraining to our high-throughput \gls{ifc} data under practical hardware constraints, we utilized distinct epoch budgets for the two datasets. For the \gls{rbc} dataset ($245,085$ images), all models were pretrained for $200$ epochs as indicated in \autoref{tab:supp:ssl_hyperparams}. Conversely, pretraining on the larger \gls{isf}\mbox{--}5CH dataset (more than $2.89$ million images) was restricted to $10$ epochs. We justify this highly compact $10$-epoch regime on the larger \gls{isf}\mbox{--}5CH dataset through both empirical stability and domain-specific visual characteristics. Unlike natural image datasets, such as ImageNet-1k, which exhibit high spatial entropy and diverse semantic classes, \gls{ifc} data are highly structured, consisting exclusively of roundish cellular morphologies on a uniform background. Under these specialized conditions, the pretraining task is primarily governed by signal learning, specifically, the extraction of micro-textures, marker localization, and signal intensity distributions across channels.

To verify the correctness and stability of the benchmarking pipeline, we use the BloodMNIST~\citep{acevedo_recognition_2019,acevedo_dataset_2020}, a publicly available dataset of the MedMNIST collection~\citep{yang_medmnist_2023}. BloodMNIST contains blood cell images annotated into eight classes (Basophil, Eosinophil, Erythroblast, Immature Granulocyte, Lymphocyte, Monocyte, Neutrophil, and Platelet). BloodMNIST is not an \gls{ifc} dataset and was acquired with the CellaVision DM96 system and consists of standard three-channel RGB images. As indicated in \autoref{tab:supp:supervised:bloodmnist}, we achieve similar performances to those from the benchmark~\citep{yang_medmnist_2023}.

\newpage

\begin{availability}
All datasets analyzed in this study are publicly available from the repositories associated with their original publications and no new datasets were generated.  All code developed for this study, covering self-supervised pretraining, benchmarking and analysis, is publicly available under the MIT License at \url{https://github.com/hciays/IFC-SSL-Bench}. The repository includes the configuration file used across all experiments, so that all results can be reproduced from the public data. Pretrained weights for the best-performing configuration identified in this study for each dataset will be deposited at Zenodo under the CC BY 4.0 license and released upon acceptance of the manuscript. The complete set of checkpoints across all method, backbone, initialization and seed combinations is not deposited owing to its size; all configurations can be reproduced from the code and configuration files in the repository, and individual checkpoints are available from the lead contact upon reasonable request. Any additional information required to reanalyze the data reported in this paper is available from the lead contact upon reasonable request.

\end{availability}

\begin{acknowledgements}
The Helmholtz Association funds this project under the research school "Helmholtz Information and Data Science School for Health (HIDSS4Health)", the program "Natural, Artificial and Cognitive Information Processing (NACIP)" and the Initiative and Networking Fund on the HAICORE@KIT partition. The state of Baden-Württemberg supports this project through bwHPC.
\end{acknowledgements}

\begin{contributions}
The authors have accepted responsibility for the entire content of this manuscript and approved its submission. We describe here the individual contributions: Conceptualization: A.J.Y.S., T.L., N.F., S.A., L.D., D.H., R.M.; Methodology: A.J.Y.S., R.-T.F., T.L., R.M.; Software: A.J.Y.S., R.-T.F.; Validation: A.J.Y.S., R.-T.F., T.L., N.F., L.K.K.S., S.A., R.M.; Investigation: A.J.Y.S., R.-T.F., T.L., N.F., L.K.K.S., S.A., R.M.; Resources: A.S., V.H., D.H., R.M.; Writing-Original Draft: A.J.Y.S., R.-T.F.; Writing-Review \& Editing: A.J.Y.S., R.-T.F., T.L., N.F., L.K.K.S., S.A., L.D., A.S., V.H., D.H., R.M.; Supervision: A.S., V.H., D.H., R.M.; Project administration: A.S., V.H., D.H., R.M.; Funding Acquisition: A.S., V.H., D.H., R.M.;
\end{contributions}

\begin{ai}
During the preparation of this manuscript, the authors used ChatGPT and Grammarly to improve the language and style. After using these, the authors reviewed and edited the content as needed and take full responsibility for the content of the manuscript.
\end{ai}

\begin{interests}
The authors declare no competing interests.
\end{interests}

\section*{References}
\bibliographystyle{theme/bxv_abbrvnat}
\bibliography{refs.bib}

\onecolumn 
\fancyhead{} 
\renewcommand{\floatpagefraction}{0.1}
\lfoot[\bSupInf]{\dAuthor}
\rfoot[\dAuthor]{\cSupInf}
\newpage

\captionsetup*{format=largeformat} 
\setcounter{figure}{0} 
\setcounter{equation}{0} 
\makeatletter 
\renewcommand{\thefigure}{S\@arabic\c@figure} 
\makeatother
\def\theequation{S\arabic{equation}}

\newpage
\section*{Framework Implementation Validation and Supervised Trainings\label{sec:supp_s1}}

{\setlength{\tabcolsep}{4pt}   
\renewcommand{\arraystretch}{0.95} 
\begin{table}[!tbh]
\centering
\caption{Accuracies of the supervised end-to-end baselines (mean $\pm$ std) on the BloodMNIST dataset. $\theta_{\mathrm{IN}}$ and $\theta_{\mathrm{Rnd}}$ denote initialization from ImageNet-supervised weights and random initialization, respectively. We report accuracies here for consistency with the MedMNIST convention~\citep{yang_medmnist_2023}. The best-performing supervised baseline for each dataset is \textbf{bolded}.
\label{tab:supp:supervised:bloodmnist}}
\begin{center}
\begin{tabular*}{240pt}{@{\extracolsep\fill}lc@{\extracolsep\fill}}
\toprule
\textbf{Backbone} & \textbf{BloodMNIST} \\
\midrule
MedMNIST ResNet50~\citep{yang_medmnist_2023}& $0.950$  \\
MedMNIST ResNet18~\citep{yang_medmnist_2023}& $0.963$  \\
ResNet50\mbox{--}$\theta_{\mathrm{Rnd}}$& $0.987 \pm 0.000$ \\
ResNet50\mbox{--}$\theta_{\mathrm{IN}}$& $0.990 \pm 0.000$  \\
\midrule
ConvNeXtV1\mbox{--}T\mbox{--}$\theta_{\mathrm{Rnd}}$& $0.947 \pm 0.000$ \\
ConvNeXtV1\mbox{--}T\mbox{--}$\theta_{\mathrm{IN}}$& $\mathbf{0.990 \pm 0.000}$ \\
ConvNeXtV2\mbox{--}T\mbox{--}$\theta_{\mathrm{Rnd}}$& $0.967 \pm 0.000$ \\
ConvNeXtV2\mbox{--}T\mbox{--}$\theta_{\mathrm{IN}}$& $0.990 \pm 0.001$ \\
\midrule
\gls{vit}\mbox{--}S/16$\mbox{--}\theta_{\mathrm{Rnd}}$& $0.955 \pm 0.000$ \\
\gls{vit}\mbox{--}S/16$\mbox{--}\theta_{\mathrm{IN}}$& $0.958 \pm 0.000$ \\

\bottomrule
\end{tabular*}
\end{center}
\end{table}
}

{\setlength{\tabcolsep}{4pt}   
\renewcommand{\arraystretch}{0.95} 
\begin{table}[!tbh]
\centering
\caption{Hyperparameters used for supervised trainings. The batch size was accordingly adjusted to eight when the training set had less then $128$ samples. "\mbox{--}" indicates unavailability.\label{tab:supp:supp_par}}
\begin{center}
\begin{tabular*}{350pt}{@{\extracolsep\fill}lcc@{\extracolsep\fill}}
\toprule
\textbf{Hyperparameter} & \textbf{Baseline } & \textbf{Finetuning} \\
\midrule
Image size & $224 \times 224$ & $224 \times 224$\\
Batch size & $128$ & $128$\\
Epochs & $200$ & $200$\\
\gls{lr} end-to-end& $1e-3$ & \mbox{--} \\
\gls{lr} backbone & \mbox{--} & $1e-4$ \\
\gls{lr} head & \mbox{--} & $1e-3$ \\
Weight decay & $1e-2$ & $1e-4$ \\
Early stopping patience & $5$ epochs & $5$ epochs \\
Optimizer & AdamW & AdamW \\
\gls{lr} scheduler & CosineAnnealingLR & CosineAnnealingLR \\

\bottomrule
\end{tabular*}
\end{center}
\end{table}
}

\begin{figure*}[!htb]
\centerline{\includegraphics[width=\linewidth]{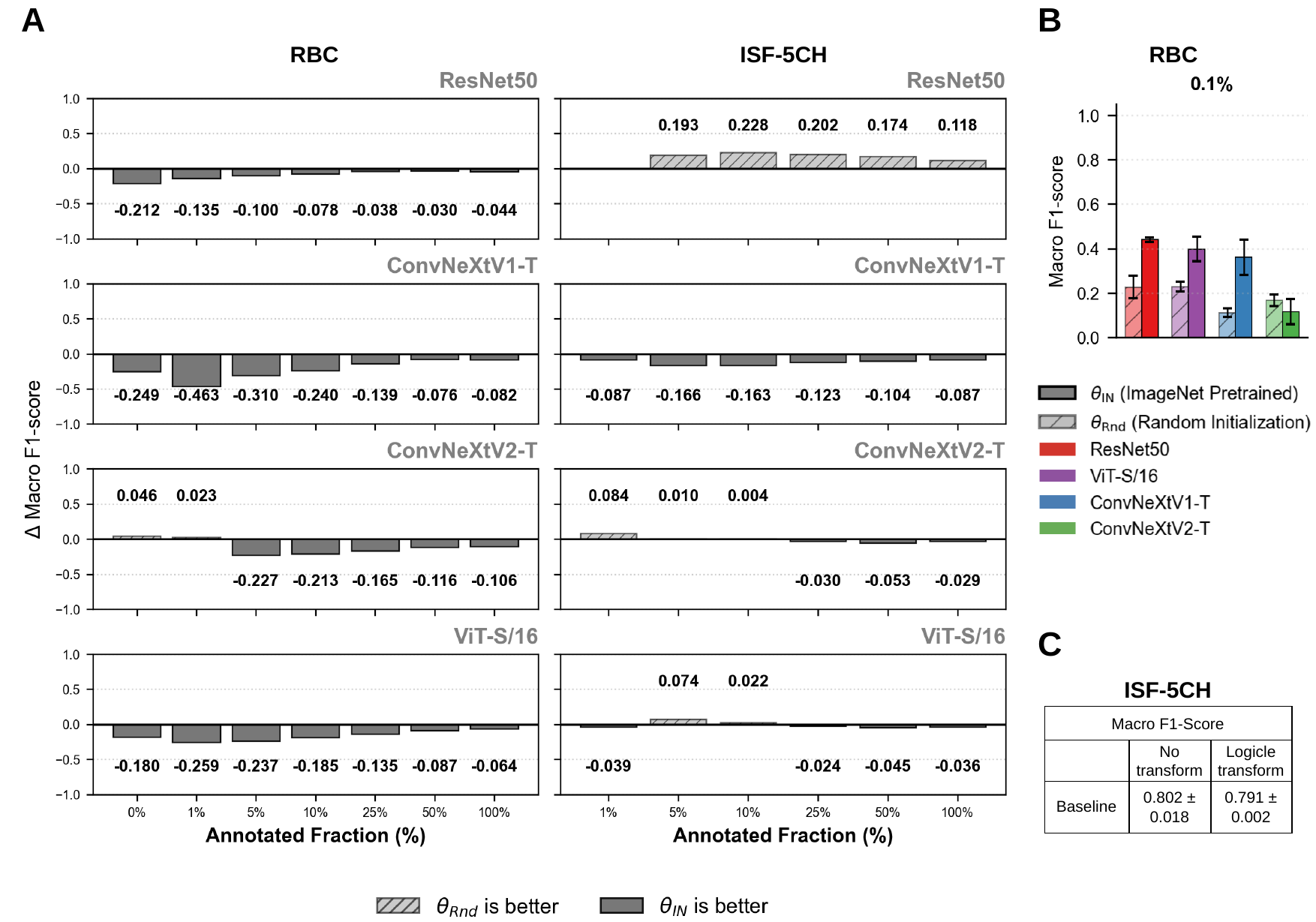}}
\caption{Supervised baseline evaluation under varying initialization, data constraints, and domain-specific transformation. \textbf{(A)} Transfer learning effect ($\Delta$ Macro F1 score = Macro F1 score $ \theta_{\mathrm{Rnd}}- $ Macro F1 score $\theta_{\mathrm{IN}}$) measured as the difference in macro F1-score between random initialization ($\theta_{\mathrm{Rnd}}$) and ImageNet-supervised ($\theta_{\mathrm{IN}}$) across different annotated fractions for the \gls{isf}\mbox{--}5CH and \gls{rbc} datasets. \textbf{(B)} Macro F1-score under extreme data scarcity ($0.1\%$ annotated fraction) on the \gls{rbc} dataset, comparing all backbone architectures across both initialization strategies. \textbf{(C)} Impact of the domain specific Logicle transformation on the \gls{isf}\mbox{--}5CH dataset. The baseline is ConvNeXtV1\mbox{--}T$\theta_{\mathrm{Rnd}}$ trained end-to-end on $100\%$ of $\mathcal{D}_{\mathrm{ann}}$. \label{fig:supp_supp_baselines}}
\end{figure*}

\section*{Self-Supervised Learning\label{sec:supp_s3}}
\subsection*{Methods\label{subsec:supp_s3:methods}}

Our analysis was focused on the following discriminative \gls{ssl} frameworks: SimCLR~\citep{DBLP:conf/icml/ChenK0H20}, MoCo-v2~\citep{DBLP:journals/corr/abs-2003-04297}, MoCo-v3~\citep{DBLP:conf/iccv/ChenXH21}, BYOL~\citep{DBLP:conf/nips/GrillSATRBDPGAP20}, Barlow Twins~\citep{DBLP:conf/icml/ZbontarJMLD21}, and DINO~\citep{DBLP:conf/iccv/CaronTMJMBJ21}. Each method was selected to evaluate different representational objectives, ranging from contrastive learning to redundancy reduction. The specific operations of these methods are summarized below. Our implementation adhered to the respective original technical specifications, preserving the primary architectural components, such as projection/prediction heads, and specialized objective functions. By utilizing a centralized configuration system for hyperparameters (e.g., temperature scaling, weight decay, and momentum coefficients), we ensured a systematic and fair comparison across different backbones and datasets. Our analysis focused on the following approaches:

\begin{itemize}
    \item \textbf{SimCLR} (Simple Framework for Contrastive Learning)~\citep{DBLP:conf/icml/ChenK0H20} is implemented as a two-view contrastive framework. It learns representations by maximizing the agreement between differently augmented views of the same image in a latent space. We utilize the normalized temperature-scaled cross-entropy loss to increase the similarity between positive pairs while minimizing agreement with other samples in the batch. 
    \item \textbf{MoCo-v2} (Momentum Contrast version 2)~\citep{DBLP:journals/corr/abs-2003-04297} utilizes a momentum-based design featuring a momentum encoder and a queue-based memory bank. This mechanism decouples the number of negative samples from the training batch size. For backbones utilizing Batch Normalization, batch shuffling is applied to maintain consistent normalization statistics between the online and momentum branches. The momentum branch is updated via an \gls{ema} of the online network, with a stop-gradient operation to ensure stable feature evolution.
    \item \textbf{MoCo-v3}~\citep{DBLP:conf/iccv/ChenXH21} builds upon the momentum-based formulation but removes the queue mechanism in favor of a purely asymmetric architecture and batch-synchronized design. In this implementation, the online branch includes an additional \gls{mlp} prediction head, while the momentum branch acts as a slowly updated teacher network. Like the previous iteration, it processes two augmented views per sample and employs an \gls{ema}-based weight update strategy to prevent representation collapse.
    \item \textbf{DINO} (Self-distillation with no labels)~\citep{DBLP:conf/iccv/CaronTMJMBJ21} employs a multi-crop self-distillation strategy. Unlike the two-view approaches above, it generates two global views and several local crops for each sample. The student network is trained to predict the outputs of a momentum-updated teacher network, which processes only the global views. A stop-gradient is applied to the teacher, which is instead updated via an \gls{ema} of the student weights, allowing the model to distill knowledge and extract consistent features across different crops without the need for negative samples.
    \item \textbf{Barlow Twins}~\citep{DBLP:conf/icml/ZbontarJMLD21} implements a redundancy-reduction objective function that aligns the representations of two augmented views while explicitly penalizing cross-correlations between embedding dimensions. By pushing the cross-correlation matrix toward the identity matrix, the method learns features that are both invariant to distortions and maximally informative.
    \item  \textbf{BYOL} (Bootstrap Your Own Latent)~\citep{DBLP:conf/nips/GrillSATRBDPGAP20} relies on an asymmetric teacher-student architecture to learn representations without the need for negative pairs. The framework minimizes the mean squared error between the online network’s prediction and the target network’s representation of the same image. The online network is equipped with a predictor head, while the target network is updated via an \gls{ema} of the online parameters, ensuring that the target provides a consistently evolving representation for the student to match.
\end{itemize}

{\setlength{\tabcolsep}{4pt}   
\renewcommand{\arraystretch}{0.95} 
\begin{table}[!htb]
\centering
\caption{List of data augmentation parameters for self-supervised pretraining. The methods using a two-view strategy are SimCLR, MoCo-v2 \& v3, BYOL, and \gls{bt}. DINO  uses a multi-crop strategy and the probability of applying Gaussian blur follows the original publication (probability of $1{.}0$ for the first global crop, $0{.}1$ for the second global crop and $0{.}5$ for all local crops). "\mbox{--}" indicates unavailability.\label{tab:supp:ssl_aug_params}}
\begin{center}
\begin{tabular*}{240pt}{@{\extracolsep\fill}lcc@{\extracolsep\fill}}
\toprule
\textbf{Augmentation technique} & \textbf{Two-view } & \textbf{DINO} \\
\midrule
Global image size & $224 \times 224$ & $224 \times 224$\\
Local image size & \mbox{--} & $96 \times 96$ \\
Number of views (global, local) & $2,0$ & $2,6$ \\
\midrule
Horizontal flip probability & $0.5$ & $0.5$ \\
Vertical flip probability & $0.5$ & $0.5$ \\
Global crop scale & $0.08 - 1$ & $0.4 - 1$ \\
Local crop scale & \mbox{--} & $0.05 - 0.4$ \\
Rotation & $0 - 180$ & $0 - 180$ \\
Translation & $0.15$ & $0.15$ \\
\midrule
Gaussian blur probability & $0.5$ & $1.0, 0.1, 0.5$ \\
Gaussian blur kernel & $23 \times 23$ & $23 \times 23$ \\
Gaussian blur sigma & $0.1 - 2$ & $0.1 - 2$ \\
Gaussian noise probability & $0.2$ & $0.2$ \\
Gaussian noise sigma & $0.1$ & $0.1$ \\
\bottomrule
\end{tabular*}
\end{center}
\end{table}
}

{\setlength{\tabcolsep}{3.5pt}   
\renewcommand{\arraystretch}{1.1} 
\begin{table}[!htb]
\centering
\caption{Hyperparameters used during self-supervised pretrainings. Common values to all methods are centralized (under MoCo-v3). "\mbox{--}" indicates unavailability. Pretraining on the \gls{isf}\mbox{--}5CH dataset occurred for only $10$ epochs.
\label{tab:supp:ssl_hyperparams}}
\begin{center}
\begin{tabular}{@{}lcccccc@{}}
\toprule
\textbf{Parameters} & \textbf{SimCLR} & \textbf{MoCo-v2} & \textbf{MoCo-v3} & \textbf{BYOL} & \textbf{\gls{bt}} & \textbf{DINO} \\
\midrule
\multicolumn{7}{l}{\textit{Common}} \\
Batch size & \multicolumn{6}{c}{$128$}\\
Embed. dim & \multicolumn{6}{c}{$256$}\\
Epochs & \multicolumn{6}{c}{$200$}\\
\midrule
\multicolumn{7}{l}{\textit{Optimization \gls{bn}}} \\
Optimizer & LARS & SGD & SGD & LARS & LARS & SGD \\
\gls{lr} & $0.15$ & $0.015$ & $0.015$ & $0.15$ & $0.15$ & $0.015$\\
Weight decay & $1e-6$ & $1e-4$ & $1e-4$ & $1e-6$ & $1e-6$ & $1e-4$\\
\midrule
\multicolumn{7}{l}{\textit{Optimization \gls{ln}}} \\
Optimizer & \multicolumn{6}{c}{AdamW} \\
\gls{lr} & \multicolumn{6}{c}{$3e-4$}\\
Weight decay & \multicolumn{6}{c}{$0.1$}\\
\midrule
\multicolumn{7}{l}{\textit{Method-Specific}} \\
Temperature & $0.2$ & $0.2$ & $0.2$ & \mbox{--} & \mbox{--} & $0.1$\\
\gls{ema} & \mbox{--} & $0.996$ & $0.996$ & $0.996$ & \mbox{--} & $0.996$\\
Memory bank & \mbox{--} & $65536$ & \mbox{--} & \mbox{--} & \mbox{--} & \mbox{--}\\
Bottleneck & \mbox{--} & \mbox{--} & \mbox{--} & \mbox{--} & \mbox{--} & $64$\\
Lambda & \mbox{--} & \mbox{--} & \mbox{--} & \mbox{--} & $0.005$ & \mbox{--}\\
\bottomrule
\end{tabular}
\end{center}
\end{table}
}
Unless stated otherwise, training hyperparameters were held constant across datasets, backbones, annotated data fractions and seeds, so that differences between methods are not confounded by differences in training budget. The complete data augmentation pipeline with its corresponding parameter ranges, and the global training hyperparameters, are given in \autoref{tab:supp:ssl_aug_params} and \autoref{tab:supp:ssl_hyperparams}, respectively. The learning rate was warmed up linearly over the first $5\%$ of optimization steps and then decayed to zero following a cosine schedule, with the total
number of steps computed as the number of epochs times the number of mini-batches per epoch.

The optimizer was the one deliberate exception, chosen to match the normalization scheme of the backbone and the reference implementation of each objective. For the \gls{bn} backbone ResNet50, LARS was used for SimCLR, BYOL and \gls{bt}, and SGD with momentum for MoCo-v2, MoCo-v3, and DINO, following the recipes published with those methods. For the \gls{ln} architectures (ConvNeXts, and \gls{vit}), AdamW was used throughout, as in the original training recipes for these backbones and in the transformer variants of MoCo-v3 and DINO. All optimiser settings are listed in \autoref{tab:supp:ssl_hyperparams}.

Some method specific settings necessarily depart from those reported in the original publications. For consistency across the benchmark and to keep the total compute tractable, we harmonized the depth and width of the projection heads, the prototype dimension of DINO, the batch size and the number of pretraining epochs rather than reproducing each method's published configuration. These values were chosen to be identical across methods wherever the objective permits, so that observed differences reflect the objectives themselves rather than unequal training budgets; the complete set of values is stated in \autoref{tab:supp:ssl_hyperparams}.

For completeness, the configuration files used to produce every reported run, together with all remaining implementation details, are available in the accompanying code repository (see data and code availability).

\newpage
\subsubsection*{Supplementary evaluation results\label{subsec:supp_s3:results}}
To provide a comprehensive overview of the interaction between architectural choice, \gls{ssl} algorithms, and initialization strategies, we present the full performance results across all experimental configurations.

\begin{figure*}[!htb]
\centerline{\includegraphics[width=1\linewidth]{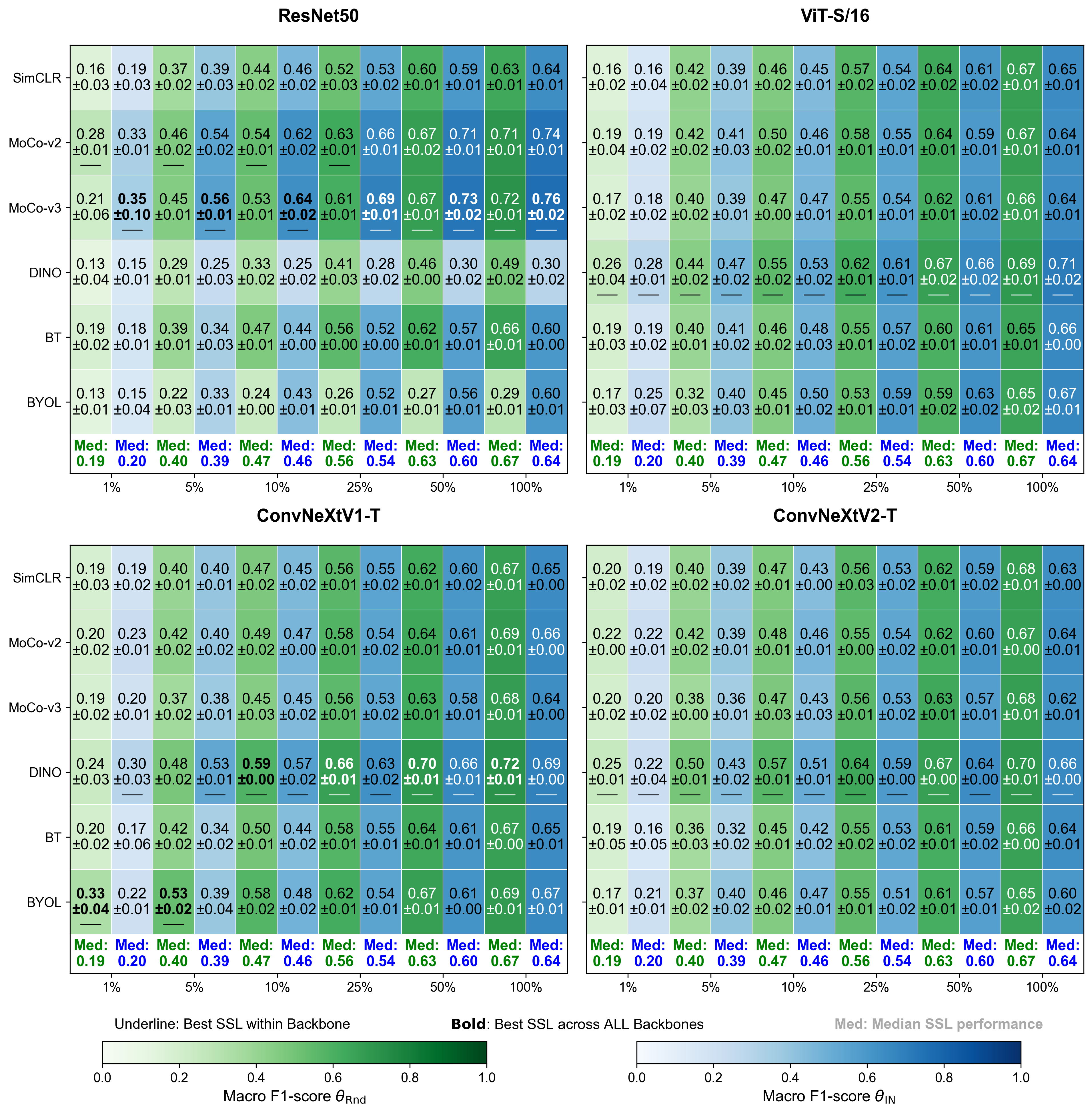}}
\caption{Comparative analysis of \gls{ssl} strategies across backbones and annotated fractions for the \gls{knn} evaluation protocol on the \gls{isf}\mbox{--}5CH dataset.
Each cell displays the test set mean macro F1-score $\pm$ standard deviation across three runs for a specific backbone, \gls{ssl} method, and weights initialization strategy. The left (green) and right (blue) halves of each cell represent random initialization ($\theta_{\mathrm{Rnd}}$) and ImageNet pretraining ($\theta_{\mathrm{IN}}$) respectively, with color intensity mapped to absolute performance $[0, 1]$. To highlight relative performance, we \underline{underline} the leading \gls{ssl} method within each specific backbone. \textbf{Bold} entries denote the globally optimal \gls{ssl} configuration across all evaluated backbones for a given data fraction and initialization mode. Bottom rows indicate the median SSL performance per fraction and initialization.  Values have been rounded to two decimal places to enhance legibility relative to \autoref{tab:res:ssl:ISF-5CH_DS:f1_knn_lp_xgb}. Furthermore, font colors within cells dynamically switch between black and white based on background intensity to ensure maximum readability of the results. \label{fig:supp_s3_isf_knn}}
\end{figure*}

\newpage
\begin{figure*}[!htb]
\centerline{\includegraphics[width=1\linewidth]{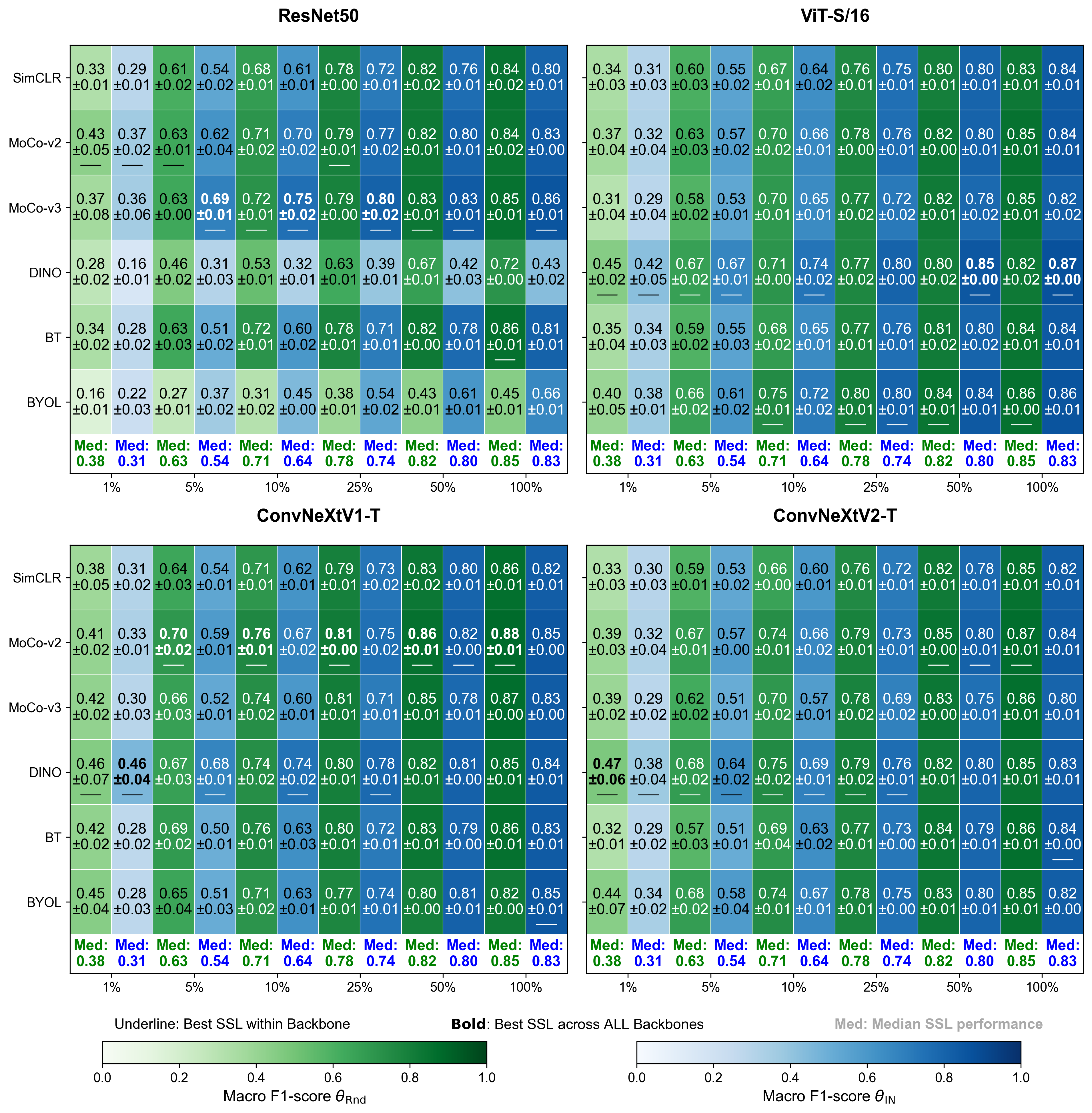}}
\caption{Comparative analysis of \gls{ssl} strategies across backbones and annotated fractions for the linear probing evaluation protocol on the \gls{isf}\mbox{--}5CH dataset.
Each cell displays the test set mean macro F1-score $\pm$ standard deviation across three runs for a specific backbone, \gls{ssl} method, and weights initialization strategy. The left (green) and right (blue) halves of each cell represent random initialization ($\theta_{\mathrm{Rnd}}$) and ImageNet pretraining ($\theta_{\mathrm{IN}}$) respectively, with color intensity mapped to absolute performance $[0, 1]$. To highlight relative performance, we \underline{underline} the leading \gls{ssl} method within each specific backbone. \textbf{Bold} entries denote the globally optimal \gls{ssl} configuration across all evaluated backbones for a given data fraction and initialization mode. Bottom rows indicate the median SSL performance per fraction and initialization.  Values have been rounded to two decimal places to enhance legibility relative to \autoref{tab:res:ssl:ISF-5CH_DS:f1_knn_lp_xgb}. Furthermore, font colors within cells dynamically switch between black and white based on background intensity to ensure maximum readability of the results. \label{fig:supp_s3_isf_lin_probe}}
\end{figure*}

\newpage
\begin{figure*}[!htb]
\centerline{\includegraphics[width=1\linewidth]{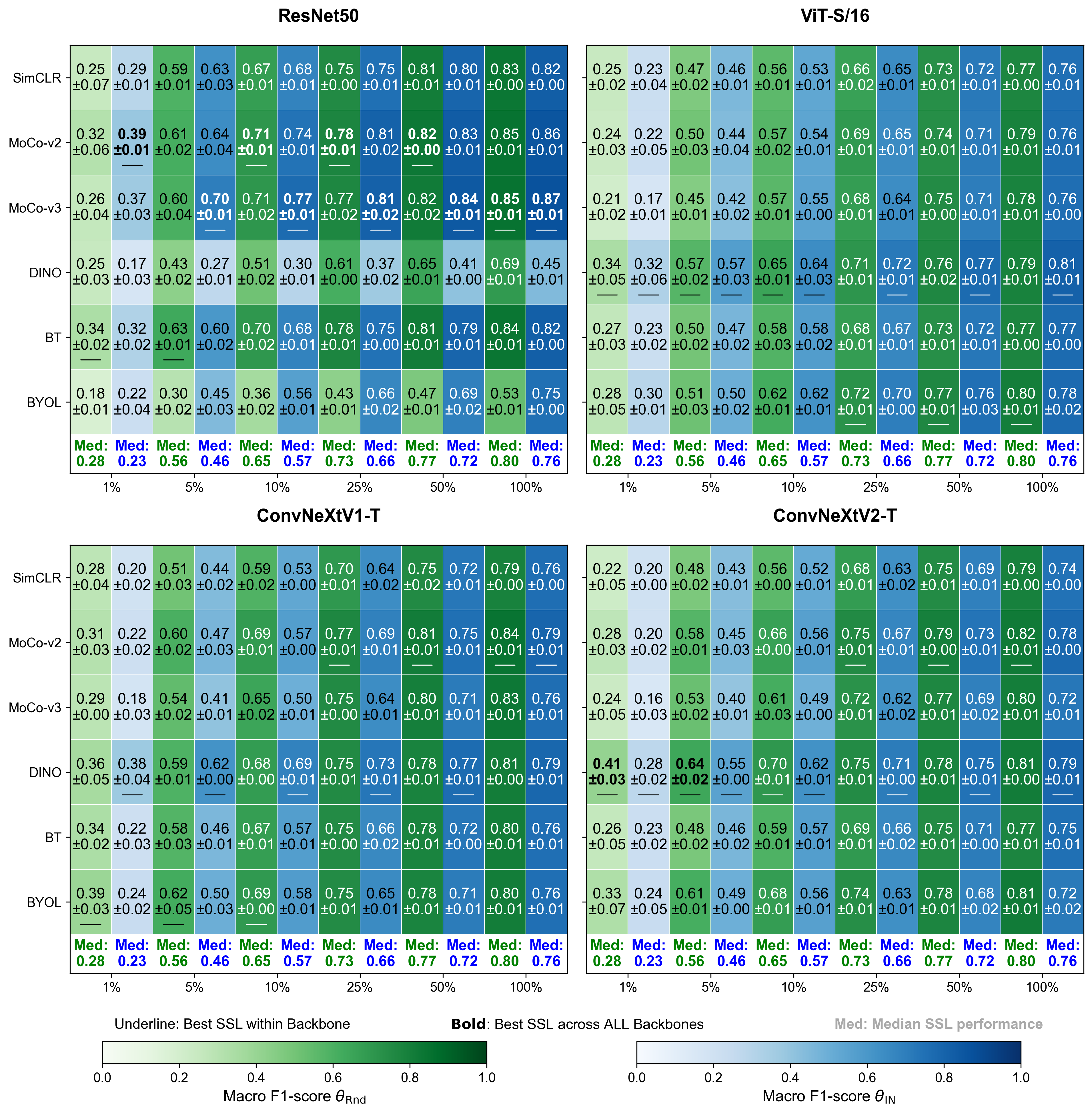}}
\caption{Comparative analysis of \gls{ssl} strategies across backbones and annotated fractions for the XGBoost evaluation protocol on the \gls{isf}\mbox{--}5CH dataset.
Each cell displays the test set mean macro F1-score $\pm$ standard deviation across three runs for a specific backbone, \gls{ssl} method, and weights initialization strategy. The left (green) and right (blue) halves of each cell represent random initialization ($\theta_{\mathrm{Rnd}}$) and ImageNet pretraining ($\theta_{\mathrm{IN}}$) respectively, with color intensity mapped to absolute performance $[0, 1]$. To highlight relative performance, we \underline{underline} the leading \gls{ssl} method within each specific backbone. \textbf{Bold} entries denote the globally optimal \gls{ssl} configuration across all evaluated backbones for a given data fraction and initialization mode. Bottom rows indicate the median SSL performance per fraction and initialization.  Values have been rounded to two decimal places to enhance legibility relative to \autoref{tab:res:ssl:ISF-5CH_DS:f1_knn_lp_xgb}. Furthermore, font colors within cells dynamically switch between black and white based on background intensity to ensure maximum readability of the results. \label{fig:supp_s3_isf_xgboost}}
\end{figure*}

\newpage
\begin{figure*}[!htb]
\centerline{\includegraphics[width=1\linewidth]{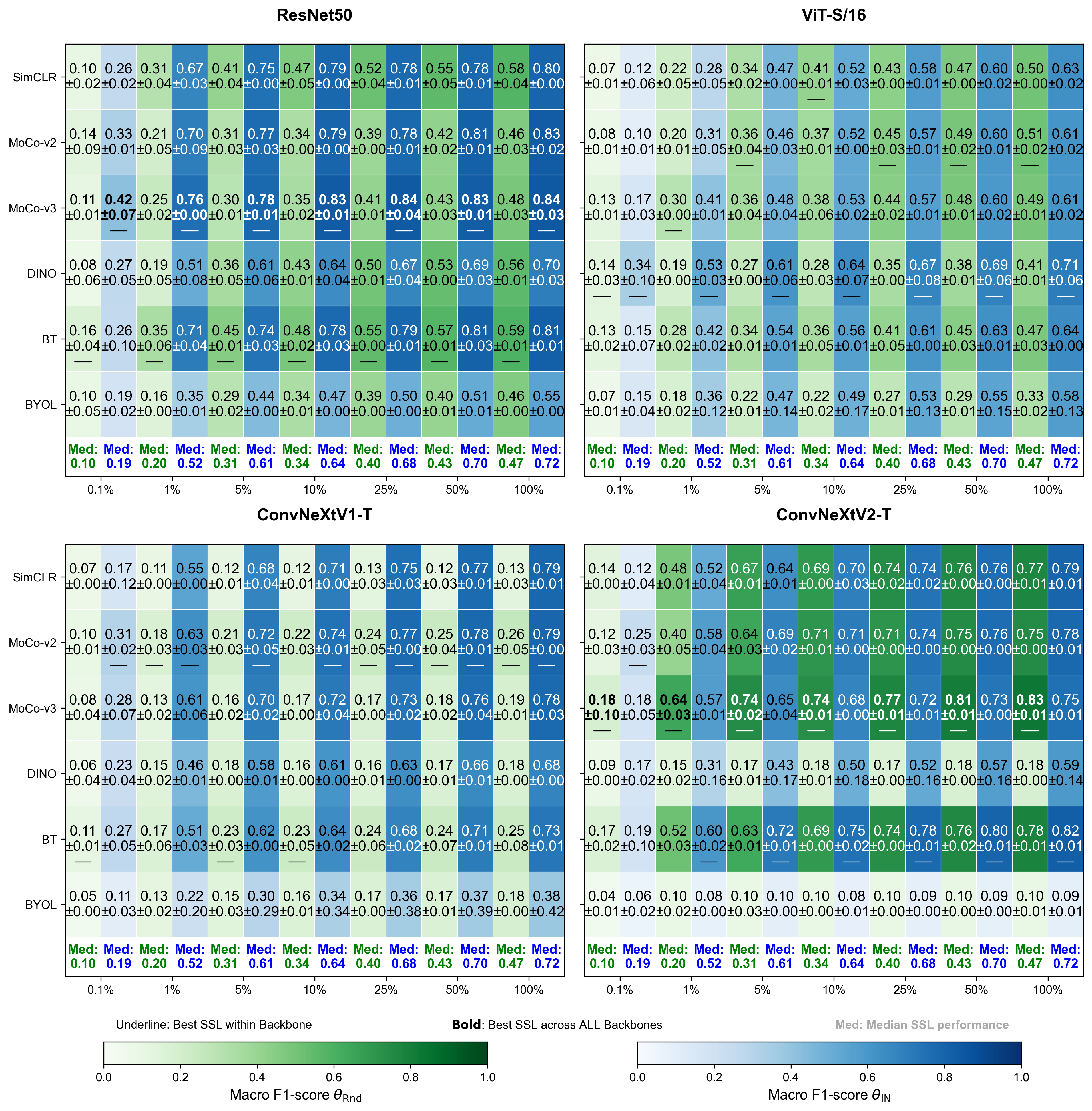}}
\caption{Comparative analysis of \gls{ssl} strategies across backbones and annotated fractions for the \gls{knn} evaluation protocol on the \gls{rbc} dataset.
Each cell displays the test set mean macro F1-score $\pm$ standard deviation across three runs for a specific backbone, \gls{ssl} method, and weights initialization strategy. The left (green) and right (blue) halves of each cell represent random initialization ($\theta_{\mathrm{Rnd}}$) and ImageNet pretraining ($\theta_{\mathrm{IN}}$) respectively, with color intensity mapped to absolute performance $[0, 1]$. To highlight relative performance, we \underline{underline} the leading \gls{ssl} method within each specific backbone. \textbf{Bold} entries denote the globally optimal \gls{ssl} configuration across all evaluated backbones for a given data fraction and initialization mode. Bottom rows indicate the median SSL performance per fraction and initialization.  Values have been rounded to two decimal places to enhance legibility relative to \autoref{tab:res:ssl:RBC_DS:f1_knn_lp_ft_xgb_split}. Furthermore, font colors within cells dynamically switch between black and white based on background intensity to ensure maximum readability of the results. \label{fig:supp_s3_rbc_knn}}
\end{figure*}

\newpage
\begin{figure*}[!htb]
\centerline{\includegraphics[width=1\linewidth]{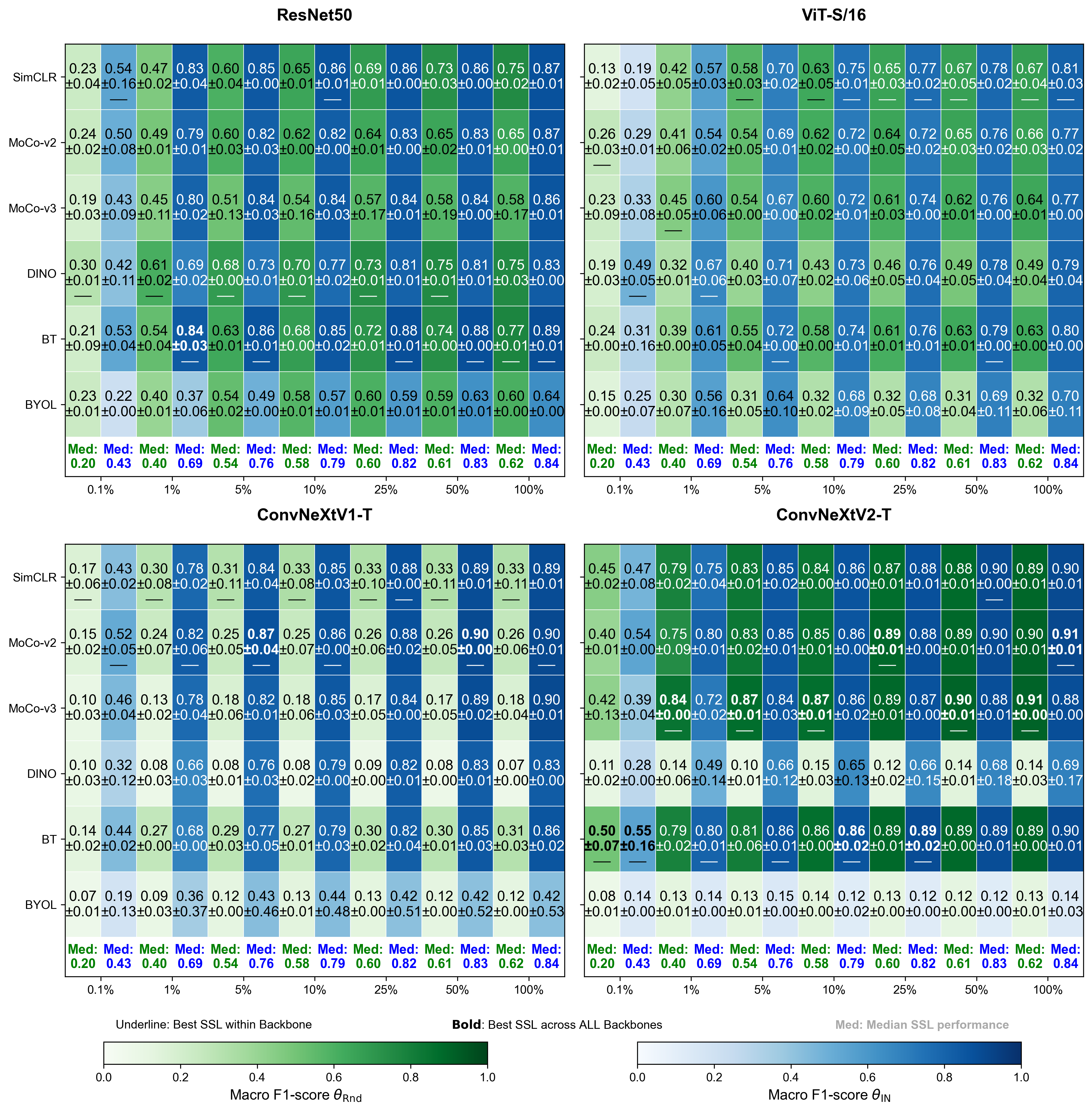}}
\caption{Comparative analysis of \gls{ssl} strategies across backbones and annotated fractions for the linear probing evaluation protocol on the \gls{rbc} dataset.
Each cell displays the test set mean macro F1-score $\pm$ standard deviation across three runs for a specific backbone, \gls{ssl} method, and weights initialization strategy. The left (green) and right (blue) halves of each cell represent random initialization ($\theta_{\mathrm{Rnd}}$) and ImageNet pretraining ($\theta_{\mathrm{IN}}$) respectively, with color intensity mapped to absolute performance $[0, 1]$. To highlight relative performance, we \underline{underline} the leading \gls{ssl} method within each specific backbone. \textbf{Bold} entries denote the globally optimal \gls{ssl} configuration across all evaluated backbones for a given data fraction and initialization mode. Bottom rows indicate the median SSL performance per fraction and initialization. Values have been rounded to two decimal places to enhance legibility relative to \autoref{tab:res:ssl:RBC_DS:f1_knn_lp_ft_xgb_split}. Furthermore, font colors within cells dynamically switch between black and white based on background intensity to ensure maximum readability of the results. \label{fig:supp_s3_rbc_lin_probe}}
\end{figure*}

\newpage
\begin{figure*}[!htb]
\centerline{\includegraphics[width=1\linewidth]{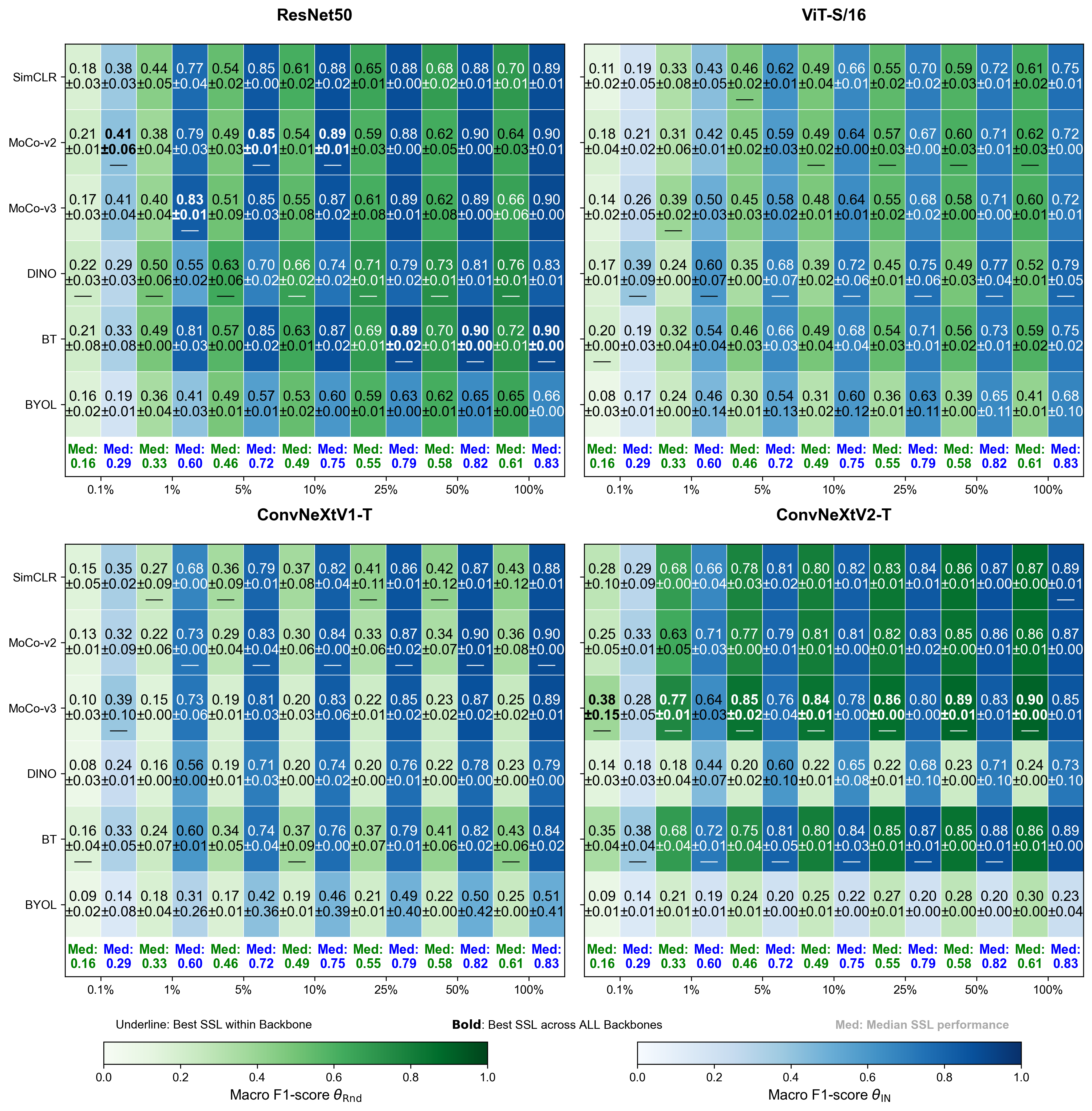}}
\caption{Comparative analysis of \gls{ssl} strategies across backbones and annotated fractions for the XGBoost evaluation protocol on the \gls{rbc} dataset.
Each cell displays the test set mean macro F1-score $\pm$ standard deviation across three runs for a specific backbone, \gls{ssl} method, and weights initialization strategy. The left (green) and right (blue) halves of each cell represent random initialization ($\theta_{\mathrm{Rnd}}$) and ImageNet pretraining ($\theta_{\mathrm{IN}}$) respectively, with color intensity mapped to absolute performance $[0, 1]$. To highlight relative performance, we \underline{underline} the leading \gls{ssl} method within each specific backbone. \textbf{Bold} entries denote the globally optimal \gls{ssl} configuration across all evaluated backbones for a given data fraction and initialization mode. Bottom rows indicate the median SSL performance per fraction and initialization.  Values have been rounded to two decimal places to enhance legibility relative to \autoref{tab:res:ssl:RBC_DS:f1_knn_lp_ft_xgb_split}. Furthermore, font colors within cells dynamically switch between black and white based on background intensity to ensure maximum readability of the results. \label{fig:supp_s3_rbc_xgboost}}
\end{figure*}

\newpage
\bibliographystyle{theme/bxv_abbrvnat}
\bibliography{refs.bib}

\end{document}